\documentclass[final,5p,times,twocolumn]{elsarticle}

\usepackage{amssymb}

\usepackage{todonotes}
\usepackage{dirtytalk}
\usepackage{rotating}           
\usepackage{multirow}
\usepackage{enumitem}
\usepackage[utf8]{inputenc}
\usepackage[T1]{fontenc}

\usepackage{amsmath}

\usepackage{subcaption}

\journal{Neural Networks}

\begin{document}

\begin{frontmatter}



\title{Continual Object Detection: A review of definitions, strategies, and challenges}


\author[inst1]{Angelo G. Menezes}
\ead{angelomenezes@usp.br}
\affiliation[inst1]{
            organization={Institute of Mathematics and Computer Sciences, University of S\~{a}o Paulo},
            addressline={Av. Trab. S\~{a}o Carlense, 400 - Centro}, 
            city={S\~{a}o Carlos},
            postcode={13566-590}, 
            state={S\~{a}o Paulo},
            country={Brazil}}

\author[inst2]{Gustavo de Moura}

\author[inst2]{Cézanne Alves}

\affiliation[inst2]{
            organization={Eldorado Research Institute},
            addressline={Av. Alan Turing, 275, Cidade Universit\'{a}ria}, 
            city={Campinas},
            postcode={13083-898}, 
            state={S\~{a}o Paulo},
            country={Brazil}}

\author[inst1]{Andr\'{e} C. P. L. F. de Carvalho}

\begin{abstract}

The field of Continual Learning investigates the ability to learn consecutive tasks without losing performance on those previously learned. Its focus has been mainly on incremental classification tasks. We believe that research in continual object detection deserves even more attention due to its vast range of applications in robotics and autonomous vehicles. This scenario is more complex than conventional classification given the occurrence of instances of classes that are unknown at the time, but can appear in subsequent tasks as a new class to be learned, resulting in missing annotations and conflicts with the background label. In this review, we analyze the current strategies proposed to tackle the problem of class-incremental object detection. Our main contributions are: (1) a short and systematic review of the methods that propose solutions to traditional incremental object detection scenarios; (2) A comprehensive evaluation of the existing approaches using a new metric to quantify the stability and plasticity of each technique in a standard way; (3) an overview of the current trends within continual object detection and a discussion of possible future research directions. 

\end{abstract}



\end{frontmatter}

\section{Introduction}
\label{sec:intro}

Deep Neural Networks (DNNs) are computational distributed models able to learn representations from raw data through a structure of hierarchical layers, similar to how
the brain handles new information. However, they are a powerful solution only when being used with data that is carefully shuffled, balanced, and standardized~\cite{hadsell2020embracing}. As real-world data may come in large streams and vary considerably from what was available during the initial training, some necessary assumptions for DNNs might not be met. In this case, they can fail entirely or suffer from a fast decay
in performance for early learned tasks, commonly described as catastrophic forgetting or catastrophic interference~\citep{robins1995catastrophic}. 

These circumstances have influenced the introduction of 
continual learning (CL),
in which techniques are mainly refined to deal with different data-dynamic scenarios. Although the interest in this area
has grown notably since
2016~\cite{parisi2019continual}, over the years several names have been used to refer to the search for models that continually adapt. Some of them are ``incremental learning'', ``lifelong learning'' and ``never-ending learning''. Yet, the recent desiderata assigned to CL models have become broader and involve not only the forgetting aspect, but also the scalability, computational efficiency, and fast adaptability features~\citep{diaz2018don}.

Within the context of computer vision, the search for strategies able to
deal with the modeling of a dynamic world is not new~\cite{ross2008incremental}. Several applications associated
with streams of images can benefit from having models able to
naturally work with changing and incremental contexts, such as autonomous cars, Unmanned Aerial Vehicles and house robots~\cite{shaheen2021continual}. Notwithstanding, most of the current
solutions for
CL 
consider the classification task as its main conundrum. In this way, the task of continual object detection, which involves both localization and classification of object samples, is not yet well 
explored, having its foundational work dating back to 2017~\cite{shmelkov2017incremental}.

Continual Object Detection (COD) is a more complex task than conventional
classification, since the predictive model
needs to deal with situations where new objects, that were unknown previously, appeared in the previous training data but were not labeled and therefore considered as ``background''. This issue affects the notion of ``objectness'' of the model and may interfere in its performance towards either favoring the detection of only previously known objects or exclusively the new ones. This tradeoff is also in part due to the natural ``tug-of-war'' effect that each task creates on the model parameters during training~\cite{hadsell2020embracing}.

This short review aims to provide an overview of the definitions, strategies, and desiderata that involve the field of COD, with the focus on exploring the scenario where object instances are introduced incrementally. To the best of our knowledge, this is the first review to address the topic and provide tools that researchers can use to standardize their research regarding incremental object detectors. In this way, we propose the following contributions:

\begin{itemize}
    \item A short and systematic review of the main strategies proposed for solving the problem of continually learning and detecting new object instances.
    \item A comprehensive evaluation of the main proposed methods for class-incremental object detection using a new metric properly adapted to identify the stability-plasticity power of a strategy according to its supposed upper-bound.
    \item An overview of the possible research directions and trends in the field.
\end{itemize}

\section{Technical Background}
\label{sec:background}

The field of continual object detection, as previously mentioned, presents the combination of CL strategies to deal with the forgetting and transferability of knowledge between object detection tasks. In this way, a general understanding of both topics is needed to identify opportunities in the field and interpret the findings of this review.

For the scope of CL, we refer to \textit{tasks} as a description of the type of prediction being made comprising a closed set of classes. For example, we can have a certain detection task $t_1$ that predicts the position and label of some classes $c_1$ and $c_2$, and another detection task $t_2$, that predicts the same for some other classes $c_3$ and $c_4$.

\subsection{Continual Learning with Neural Networks}
\label{sec:cl-background}

Continual learning, or lifelong learning, has been coined as the ability to learn consecutive tasks without forgetting how to perform on the previously trained ones~\citep{thrun1995lifelong}. Some researchers have pointed out over the years that research on this topic might lead to the development of an artificial general intelligence~\citep{silver2011machine, clune2019ai} since such behavior is expected from intelligent agents.

As the amount of data available increases over the years and current machine learning (ML) systems still have poor ability to solve for new tasks without being properly retrained, solutions that involve continual and multi-task learning will become more prevalent~\citep{rebuffi2017icarl}. Also, as deep learning techniques are the state-of-the-art for several tasks in areas such as computer vision and natural language processing~\citep{ren2015faster, devlin2018bert}, the adaptation of the ongoing strategies in these fields for the continual paradigm becomes a natural promising research direction.

Despite not being a new research topic~\citep{thrun1995lifelong}, there is still no consensus on all the characteristics that a CL model should consider essential (i.e., CL Desiderata) during its optimization process~\citep{aljundi2019continual, mundt2020wholistic}. Most of the definitions favor a specific direction based on the researched topic the author is involved. For example, one may say that constant memory and forward transfer are fundamental for robotics. At the same time, for recommendation systems, one could argue that online learning and fast adaptation are more important features. Following this line of thought, for the continual object detection venue, in especial the class-incremental setting, we argue that the following desiderata should be aimed:  

\begin{itemize}
    \item \textbf{Quasi-constant memory}: A CL model should work with bounded memory.
    \item \textbf{Backward Transfer}: A CL model should have the ability to improve the performance of previously learned tasks by learning a new one.
    \item \textbf{Forward Transfer}: A CL model should have the ability to improve the performance of future tasks using previously acquired knowledge.
    \item \textbf{Fast adaptation and recovery}: A CL model should be able to adapt quickly for new tasks, and, in case a class was gracefully forgotten (better described in the work of ~\citet{ahn2019uncertainty}, the model should recover the previous performance at the same speed.
\end{itemize}

Also, the ability to identify when a sample object is unknown at test time and decide whether to learn from it during incremental training is of interest for applications in autonomous robots~\citep{joseph2021towards}. This scenario, which is related to other different ML paradigms (e.g., out-of-distribution detection, open-set and open-world recognition), might be a pursued direction for having less human interference in the learning process~\citep{mundt2020wholistic, mundt2021cleva}.

\subsubsection{Scenarios}
When working with classical CL benchmarks~\citep{lomonaco2017core50}, there are three general situations in which data might be introduced:
\begin{itemize}
    \item \textbf{New Instances (NI)}: New training samples of previously known classes.
    \item \textbf{New Classes (NC)}: Only new training samples of new classes.
    \item \textbf{New Instances and Classes (NIC)}: New training samples from both old and new classes.
\end{itemize}

When working in classification tasks, the presence of the task ID dictates the space of possible classes and distributions that can be recognized during test time. Thus, it describes whether it is possible to create task-specific solutions or if a more general CL strategy is needed~\citep{delange2021continual}. Following this trend, the CL literature has mostly adopted the convention from~\citet{van2019three} for three general task scenarios:

\begin{itemize}
    \item \textbf{Task-Incremental Learning}: Assumes the model has information about the task ID during training and testing. The situation allows for task-specific solutions. 
    \item \textbf{Domain-Incremental Learning}: Assumes the task ID is not given during test time, but the structure of the task is maintained. Class labels are usually kept, but the data distribution might change. 
    \item \textbf{Class-Incremental Learning}: Assumes the task ID is not given during test time, and model needs to infer it. In this way, the model needs to expand its range of predictions and incrementally add new classes.
\end{itemize}

\begin{figure}[!htb]
\centering
\includegraphics[width=\linewidth]{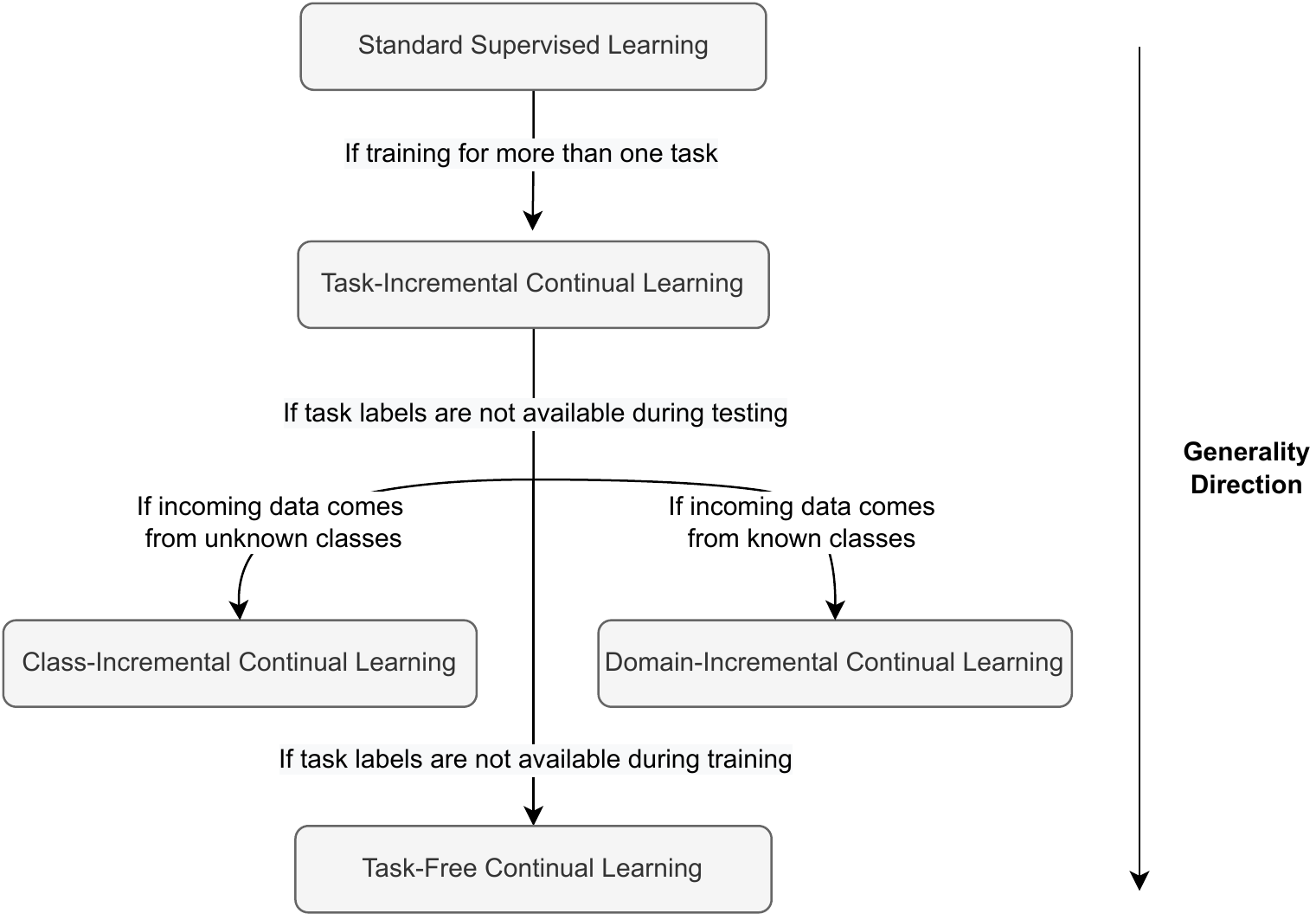}
\caption{General scenarios for CL}
\label{fig-CL-scenarios}
\end{figure}

Additionally, Task-Free or Task-Agnostic CL~\citep{aljundi2019task, normandin2021sequoia} represents an additional scenario for when the task labels are not given during either training or testing, which makes it the most challenging scheme. For that, the model does not have any information on task boundaries and still needs to deal with data distribution changes. The generality related to each mentioned scenario is described in Figure~\ref{fig-CL-scenarios}. 

\subsubsection{Evaluation}

For evaluating CL models on incremental benchmarks, metrics should assess the desired characteristics we expect the system to have. To this extent, a CL model, in general, should be evaluated not only on its final performance but also on how transferable its knowledge is and how fast it learns and forgets tasks. The usual procedure adopted by the CL community to comply with this scheme was first introduced by ~\citet{lopez2017gradient} with three metrics. 

Average Accuracy (ACC) is the average final accuracy over all seen $T$ tasks as described by Equation~\ref{eq-acc}.  

\begin{equation}
    ACC = \frac{1}{T} \sum_{i=1}^{T}R_{T,i}
    \label{eq-acc}
\end{equation}

Backward Transfer (BWT), as shown by Equation~\ref{eq-bwt}, is the measure of the influence that learning a new task has on the tasks learned so far. A negative value for this metric indicates the forgetting of old classes.

\begin{equation}
    BWT = \frac{1}{T-1} \sum_{i=1}^{T-1}R_{T,i} - R_{i,i}
    \label{eq-bwt}
\end{equation}

Forward Transfer (FWT), as demonstrated in Equation~\ref{eq-fwt}, represents the impact that learning a new task will have on the consecutive tasks. A positive forward transfer is an indication that the model can perform ``zero-shot'' learning. 

\begin{equation}
    FWT = \frac{1}{T-1} \sum_{i=2}^{T}R_{i-1,i} - \bar{b}_{i}
    \label{eq-fwt}
\end{equation}

For these metrics, $R_{i,j}$ stands for the final test accuracy on task $t_{j}$ after observing the samples of task $t_{i}$, and $\bar{b}$ the test accuracy of each task when trained with random initialization. The metrics above assume the model has access to all tasks beforehand and can be evaluated on all $T$ tasks right after it finishes the training in each individual task $t_{i}$. 

For measuring how far an incremental model response is from an ideal setting and therefore assessing its overall stability-plasticity, \citet{hayes2018new} proposed $\Omega$ as the ratio between the model`s response and the one from the joint-training equivalent (i.e., a model trained offline with all task data) as shown by Equation~\ref{eq-ratio}. We will refer to this metric as the upper-bound ratio.

\begin{equation}
    Upper\mbox{-}bound\ ratio \ (\Omega_{all}) = \sum_{t=1}^{T}\frac{R_{T,t}}{R{joint,t}}
    \label{eq-ratio}
\end{equation}

Although there are interesting adaptations of these metrics that account for the performance of a CL model along each timestep in training time, in an application context, a good final performance at test time is usually what is considered. Additionally, some other metrics provide helpful information regarding the CL desiderata, such as computational efficiency and memory size~\citep{diaz2018don}, but we will not explore them in the current context of this review.

\subsubsection{Strategies}

Research to overcome catastrophic forgetting is as old as the own field of neural networks~\citep{rumelhart1992reducing, robins1995catastrophic}, but previously had its focus on solving the problem for shallow networks. When dealing with deep architectures, the main methods have been commonly divided into three families of techniques based on: parameter isolation, regularization, and replay~\citep{delange2021continual}.

\textbf{Parameter isolation techniques}

Parameter isolation strategies aim to mitigate forgetting by specifying parameters to deal with each individual task. This setup typically requires the freezing of some network parameters and then either dynamically expanding the network`s capacity~\citep{yoon2017lifelong} when new tasks arrive or learning specific sparse masks~\citep{mallya2018piggyback}. One of the base works for this family was proposed by \citet{rusu2016progressive}, where a deep neural network column of layers is trained to execute a single task. When a new task arrives, the previously trained weights are frozen, and a new column of layers with a lateral connection to the first column is added and then trained to execute the new task. Other works also expand on this strategy to deal with the issues caused by the increased final model size by applying network pruning and quantization~\citep{hung2019compacting}. For this family of techniques, it is generally guaranteed that the network will perform equally well as if it was trained from scratch at the cost of having a more significant memory footprint. Additionally, models in this group often have the disadvantage of needing a task oracle to reveal the task ID at test time~\citep{delange2021continual}.

\textbf{Regularization-based techniques}

Regularization-based methods introduce strategies to prevent the network parameters from deviating too much from the learned values that performed well for the old classes. The Elastic Weight Consolidation (EWC) strategy proposed by ~\citet{kirkpatrick2017overcoming} first finds important parameters for the learned tasks and then penalizes their changes when new tasks are presented. Besides penalty-based regularization, ~\citet{li2017learning} proposed the Learning Without Forgetting (LWF) strategy in which a copy of the network trained on the base classes is created and knowledge distillation is applied to transfer the knowledge of the copy to the network trained on the new data. For this whole family of methods, there is generally no need for storing old data or changing the current architecture. This is based on the assumption that the task`s knowledge is included on the weights, and can be preserved by either penalizing their change directly or by constraining the updates for new data using the old activations and logits. However, for this group of techniques, performance is often limited when compared to other CL strategies~\citep{pellegrini2019latent, beaulieu2020learning}.  

\textbf{Replay techniques}

Methods based on replay, often called rehearsal, store samples from previously seen data or use generative models to create pseudo-samples that follow the previous data distribution. The replay samples are then mixed with the ones of the new task to ensure that the data distribution of the new task does not deviate much from the previously learned data distribution. Following this line,~\citet{rebuffi2017icarl} proposed the iCaRL strategy in which the samples that best represent the class means in the feature space are stored and used at test time with a nearest-mean classifier. In a different way,~\citet{lopez2017gradient} proposed the Gradient Episodic Memory (GEM) technique to constrain the model optimization by using replay samples to limit the gradients for the new task in a way that the approximated loss from the previous tasks will not increase.  

When working with unstructured data (e.g., images and videos), the required memory buffer to store old samples might be considerably large, making its use impracticable for some real-world scenarios~\citep{pellegrini2019latent}. Techniques based on pseudo-rehearsal, a.k.a. generative replay, were established to overcome this limitation.~\citet{shin2017continual} proposed to train a generative model on the old data distribution and use it to generate fake samples that help in mitigating the forgetting of old classes. Although having the downside of the model`s performance being upper-bounded by the joint-training in all tasks~\citep{delange2021continual}, the replay family has been the most consistently used strategy in real-world applications of CL~\citep{shaheen2021continual, shieh2020continual}.


\subsubsection{Other Continual Learning Paradigms}

Some other learning paradigms have been adjusted to diminish the forgetting of CL systems by allowing the model to learn the desired adaptability and stability directly from the data~\citep{caccia2021special, hospedales2020meta}.

\textbf{Meta-Learning for Continual Learning}

Meta-learning, a.k.a. ``learning-to-learn'', uses knowledge obtained from learning tasks to improve the learning of new ones. Because of the general terminology, there are several perspectives proposed in the literature that relate to the topic, such as transfer learning, AutoML, and multi-task learning~\citep{hospedales2020meta}. In the context of neural networks, meta-learning has been framed as an end-to-end pipeline with two levels where an outer algorithm adjusts the learning of an inner algorithm so that the outer model objective is improved in the end. In simpler words, it is the search for inductive biases in a neural network that leads to the fulfillment of a meta-level objective. This meta-objective can be applied for diverse goals such as generalization performance, fast adaptation, or even the avoidance of catastrophic forgetting~\citep{flennerhag2019meta}.  

The application of meta-learning to solve CL meta-objectives has been referred to as meta-continual learning (Meta-CL)~\citep{caccia2020online} and can take different forms.~\citet{rajasegaran2020itaml} introduced the use of meta-learning for finding a set of generic weights that can generalize well for all seen tasks by quickly adapting to them at test time with minimum forgetting.~\citet{javed2019meta} proposed a meta-objective for finding task-independent network representations that minimize the forgetting of old tasks and accelerate future learning of new ones.~\citet{beaulieu2020learning} presented the ANML strategy which uses a neuromodulatory network to modulate the learning of a base network by gating the neurons in a specific layer during the forward and backward passes. 

\textbf{Self-Supervision for Continual Learning}

Self-supervision is the paradigm in which the data generates its own labels and learns to predict them back as a pretext task. Some examples of pretext tasks are colorizing grayscale images, predicting rotation of objects, and matching different augmented views of the same image~\citep{huang2021survey}. The advantage of having the data to generate its own supervision signal is to be able to use large-scale unlabeled datasets and obtain robust representations that can be used for other downstream tasks such as image classification, object detection, and semantic segmentation~\citep{jing2020self}. Recently, self-supervised pre-trained networks outperformed their supervised counterpart for downstream tasks of classification and detection in large benchmarks~\citep{caron2020unsupervised, bar2021detreg}.

In the context of CL, the feature extraction backbone is generally frozen for not allowing gradual changes in the representations during online updates. This inevitably causes the need for networks that can produce more general features, which favors the use of self-supervision in their training. In fact,~\citet{gallardo2021self} showed empirically that self-supervised pre-trained models provide representations that generalize better for class-incremental learning scenarios.~\citet{pham2021dualnet} proposed a learning structure based on the human brain complementary learning system, in which a model is optimized via self-supervision on stored samples to produce general representations that are then refined by supervised learning for quick knowledge acquisition on the labeled data. Beyond that, ~\citet{caccia2021special} expanded the generality of self-supervised representations to the meta-learning world by having models optimized to match different augmented views of the same image and at the same time generate representations that minimize the forgetting of old classes.

\subsection{Object Detection with Neural Networks}

Object detection is a computer vision task that involves the localization and classification of items of interest in an image. The goal of an object detector is to predict the coordinates of each bounding box that surrounds the objects of interest and assign a category to it. Previous to 2012, most solutions related to the topic were based on heuristics and hand-crafted visual descriptors~\citep{viola2001rapid, lowe2004distinctive} which limited its application in several domains. After the success that convolutional neural networks (CNNs) had in generating rich features for classification, they started to compose strategies for the more challenging task of object localization and recognition~\citep{girshick2014rich, girshick2015fast}. Since then, they have presented outstanding results in large competitions related to the detection task and became their baseline solution~\citep{wu2020recent}.

Object detectors based on DNNs can usually be divided into two modalities: two-stage and one-stage detectors. Both have in common the presence of a backbone network for providing useful feature maps to be used in localization and identification of object categories~\citep{huang2021survey}. These features can be resumed in a single 3D tensor extracted directly from the output of a single layer in a pre-trained architecture (e.g., C4 layer in ResNet-50) or a multi-dimensional tensor resulting from the gathering of the output of several layers from a top-down architecture with lateral pathways as in the work of~\citet{lin2017feature}. The backbones used for detection tasks are generally deep CNNs pre-trained on large image datasets (e.g., ImageNet) intended for classification~\citep{deng2009imagenet}.

\subsubsection{Two-Stage Detectors}

\begin{figure}[h]
\centering
\includegraphics[width=\linewidth]{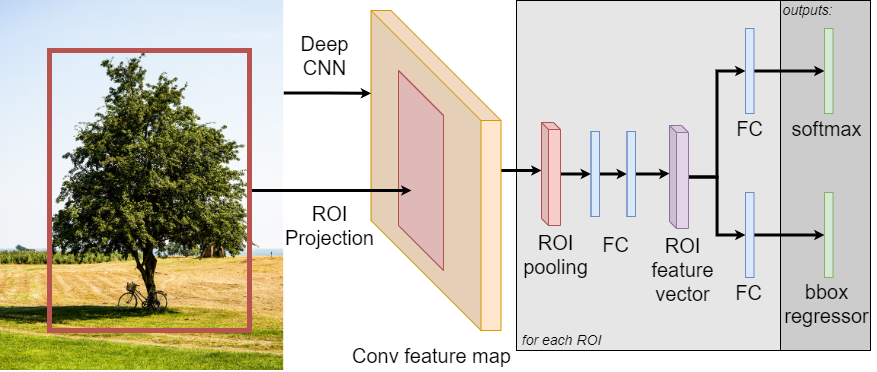}
\caption{Fast-RCNN architecture.}
\label{fig-Fast-RCNN}
\end{figure}

This class of detectors uses a separate structure to generate a set of ``guesses'' of where the objects are present in the image. These assumptions on the image, also called region proposals or just proposals, will be then classified into the known categories and have their bounding box refined to correctly identify the object's limits. R-CNNs~\citep{girshick2014rich} were one of the first two-stage strategies for object detection and used Selective Search~\citep{uijlings2013selective} for selecting its region proposals. The problem with this setup was that every proposal was processed separately by the CNN for feature extraction, which caused the inference process to be too slow. In the following work of the same authors, they propose the Fast-RCNN~\citep{girshick2015fast} in which a CNN first processes the image to extract the features maps. Then, the external proposals are used to select the regions within the feature maps through a Region of Interest (RoI) pooling layer, to be processed by the classification and regression heads as illustrated in Figure~\ref{fig-Fast-RCNN}. 

In the work of~\citet{ren2015faster}, the authors ceased the use of heuristics for selecting region proposals by using a separate network called Region Proposal Network (RPN) able to be optimized specifically for identifying more probable regions of objects within an image. Their solution used the same structure as Fast-RCNN. Still, it was way faster than its counterpart, which resulted in it being named Faster-RCNN.~\citet{lin2017feature} improved the network backbone performance in generating robust features able to identify smaller objects. Their strategy, called Feature Pyramid Networks (FPN), exploited the ``inherent multi-scale pyramidal hierarchy'' that deep CNNs carry through exploring a top-down architecture with lateral connections that helps in the propagation of information from the higher layers to the lower ones. 


\subsubsection{One-Stage Detectors}

One-stage models, also known as single-stage detectors, are often faster than their two-stage counterparts at the cost of having lower predictive performance~\citep{huang2021survey}. There is no region proposal heuristic or network for this class of models since it usually considers that every position on the image might have an object, leaving the model to classify each position as either background or the target category. The You Only Look Once (YOLO) detector~\citep{redmon2016you} was one of the first successful models to show a good balance between accuracy and speed by dividing the whole image into a set of grid cells and predicting the presence of one or more objects in each of them. 

\begin{figure}[!ht]
    \centering
        \begin{subfigure}[b]{\columnwidth}
            \centering
            \includegraphics[width=\columnwidth]{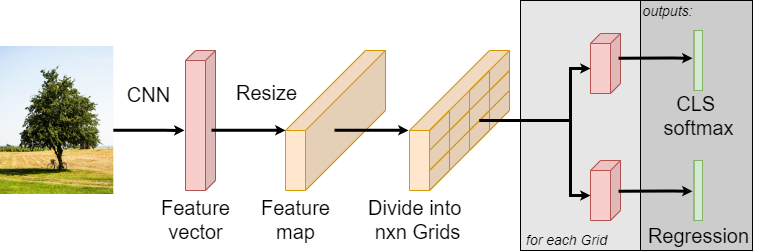}
            \caption{YOLO}
            \label{fig:yolo}
        \end{subfigure}
    \hfill \\
        \begin{subfigure}[b]{\columnwidth}
            \centering
            \includegraphics[width=\columnwidth]{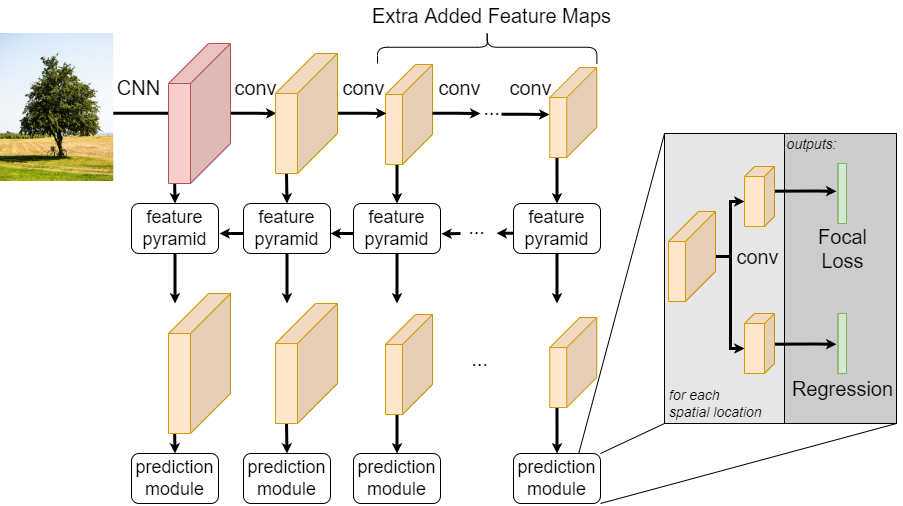}
            \caption{RetinaNet}
            \label{fig:retinanet}
        \end{subfigure}
    \caption{The general pipeline throughout the YOLO and RetinaNet architectures.}
    \label{fig-One-Stage}
\end{figure}

Improving on the inferior ability of the first YOLO architecture for detecting smaller objects, \citet{liu2016ssd} proposed the Single Shot Multibox Detector (SSD), which made use of a more elaborated CNN architecture and a set of pre-defined anchors in multiple scales and aspect-ratios. These additional features helped the model reach a decent performance while still operating in real-time. Building on top of that, \citet{lin2017focal} RetinaNet focused on dealing with the large number of negative samples that are generated by the pre-defined anchors using their Focal Loss, which weights down the importance of easy negative samples while increasing the focus of the network weight updates on the hard ones. This network also uses FPN in its architecture and has reached results that compare to Faster-RCNN. An illustration of the general pipeline used in the YOLO and RetinaNet detectors is shown in Figure~\ref{fig-One-Stage}.


Later on, several versions of the YOLO architecture, which are commonly referred to as the ``YOLO family'', have been proposed and optimized for decreasing the gap against two-stage models regarding $mAP$ performance~\citep{ge2021yolox} while keeping the real-time characteristic. Moreover, recently a few more elaborated strategies, such as CenterNet~\citep{duan2019centernet} and FCOS~\citep{tian2020fcos}, that do not make use of either pre-defined anchor boxes or proposals, have raised the bar for the performance in popular detection benchmarks.

\subsubsection{Benchmarks}

Training large DNNs requires the availability of large datasets since they tend to be more accurate as more data gets processed~\citep{lecun2015deep}. Considering that detection annotations are harder to be obtained than just labels for the whole image, the most popular benchmarks on the topic have become the ones from competitions organized by resourceful universities or big tech companies. The two most explored are the Pascal VOC~\citep{everingham2010pascal} and MS COCO ~\citep{lin2014microsoft}. Although there are different versions of the datasets based on the year of the challenges, researchers have adopted the VOC 2007 and COCO 2014 as references. Table~\ref{tab-benchmarks} displays some statistics related to these benchmarks.

\begin{table}[!h]
    \centering
    \caption{Statistics for the main object detection benchmarks~\citep{zou2019object}.}~\label{tab-benchmarks}
    \scalebox{0.75}{
    \begin{tabular}{lrr}
    \hline
    
    Dataset                                                    & VOC 2007 & COCO 2014  \\ 
    \hline
    Number of classes                                                                       & 20                 & 80                  \\
    Number of training images (train+val)                                                   & 5,011              & 123,287              \\
    Number of training instances                                                            & 12,608             & 896,782              \\
    Number of testing images                                                                & 4,952              & 81,434               \\
    Mean of bounding boxes per each training image                                          & 2.51               & 7.27              \\
    \hline
    \end{tabular}}
    \end{table}

Recently, the LVIS dataset~\citep{gupta2019lvis} was released with the promise of being a more complex (and natural) benchmark due to its vast number of categories but a low amount of samples in some of them. The dataset has over 164,000 images with more than 1000 categories and 2.2 million high-quality annotations, making it a tough challenge for generalization on ``long-tailed'' categories.

\subsubsection{Evaluation}

\begin{figure}[h]
\centering
\includegraphics[width=.7\linewidth]{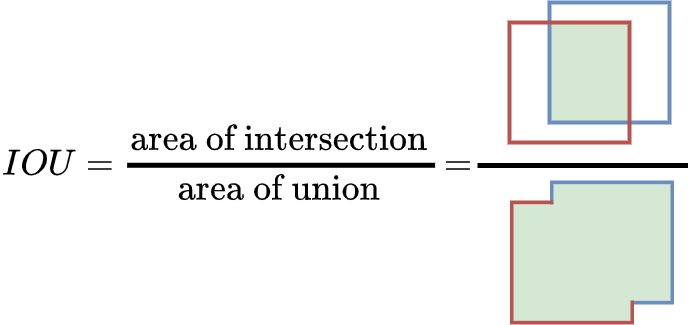}
\caption{Illustration of the Intersection Over Union equation. Image adapted from~\citet{padilla2020survey}.}
\label{fig-IOU}
\end{figure}

The evaluation of object detection models is conducted by assessing how much each predicted bounding box misses or hits a ground truth based on a threshold. The equation that governs this metric is the intersection over union (IOU), also known as the Jaccard Index (Equation~\ref{eq-iou}) in which $B_{pred}$ is the coordinate of the predicted bounding box and $B_{gt}$ is the ground truth equivalent~\citep{padilla2020survey}. An illustration of these terms is shown in Figure~\ref{fig-IOU}. 

\begin{equation}
    Jaccard \ Index = IOU =\frac{area(B_{pred} \cap B_{gt})}{area(B_{pred} \cup B_{gt} )}
    \label{eq-iou}
\end{equation}

The threshold value indicates how much overlap is needed to consider that a prediction was in fact a true positive. Then, the comparison of detection models can be made by calculating the average precision (AP) (i.e., the ratio of true positives over the sum of true positives and false positives) and average recall (AR) (i.e., the ratio of true positives over the sum of true positives and false negatives) for a given threshold. Equations~\ref{eq-prec} and~\ref{eq-recall} describe both metrics. 

\begin{equation}
    Precision = \frac{TP}{TP + FP} = \frac{TP}{all \ detections}
    \label{eq-prec}
\end{equation}
\begin{equation}
    Recall = \frac{TP}{TP + FN} = \frac{TP}{all \ ground \ truths}
    \label{eq-recall}
\end{equation}

A common value used for the threshold is $0.5$ (e.g.: $AP^{50}$). The standard evaluation procedure is to consider the mean average precision ($mAP$) at a given threshold for all classes that a detector is able to recognize. Moreover, for better dealing with false negatives, the $mAP$ term is commonly assigned as the area under the curve (AUC) of the precision against the recall curve using the specified threshold~\citep{padilla2020survey}. In addition to that determination, for some situations, benchmarks may also use the mean over the average precision of each class for several thresholds (e.g., $mAP@[.5:.95]$)~\citep{lin2014microsoft} to indicate a more stable performance.


\section{Continual Learning for Object Detection}
\label{sec:COD}

\begin{figure*}[!htb]
	\begin{center}
		\includegraphics[width=1.5\columnwidth]{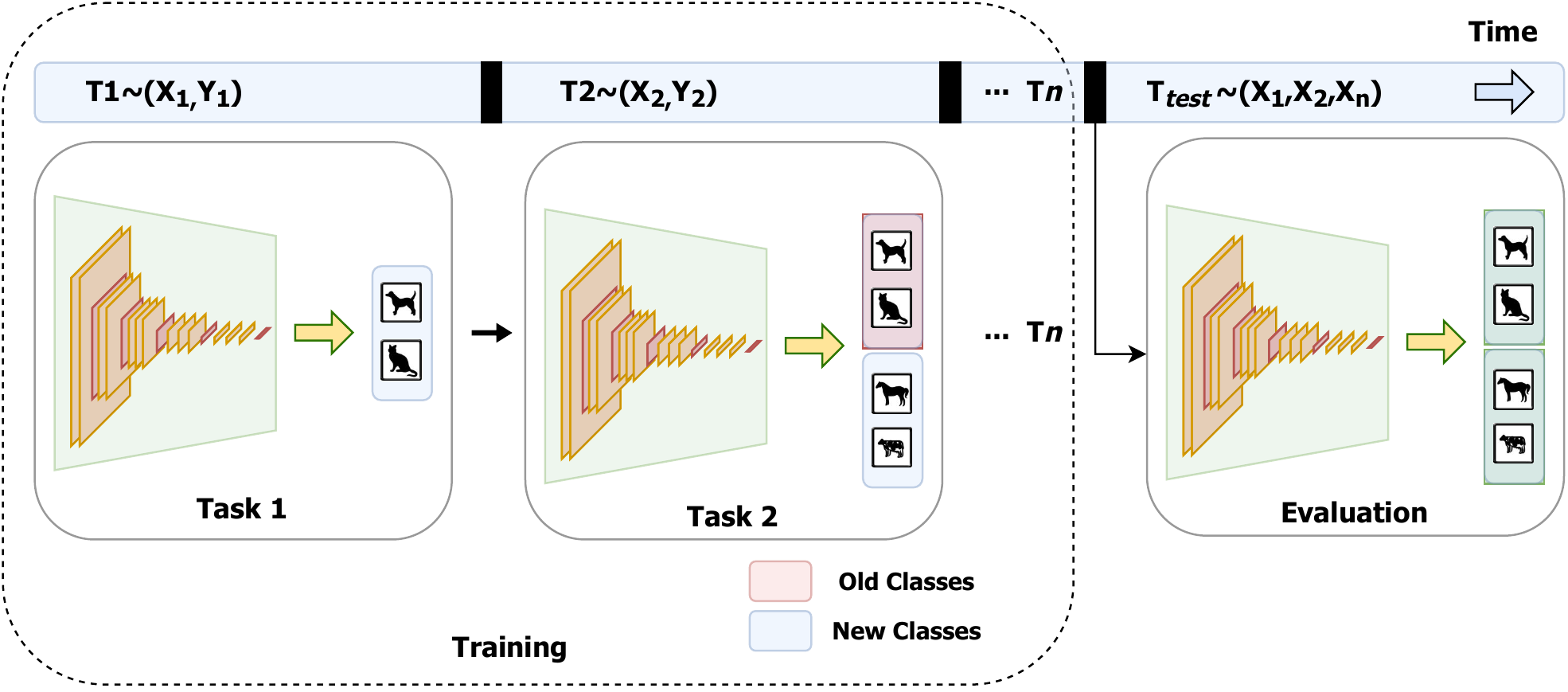}
		\caption{A generic class-incremental scenario for object detection.}
        \label{fig:cil_desc}
	\end{center}
\end{figure*}

The general goal of the continual learning paradigm for object detection is to learn a sequence of tasks $[t_1, t_2, t_3, ...]$ and have a model able to successfully localize and identify all the involved classes at test time as illustrated by Figure~\ref{fig:cil_desc}.

The area of applications of continual learning methods to the object detection task is still young and in active development~\citep{shmelkov2017incremental, peng2020faster, yang2021multi}. Strategies are mostly split into two large pools: Class-Incremental Object Detection (CIOD) and Domain-Incremental Object Detection (DIOD). The former looks at problems where the model has learned the representation of base classes and then needs to extend its prediction power over new unknown classes sequentially. The latter is formed by solutions to problems where the classes are fixed, but their distribution can change over time. In this situation, the model needs to be able to identify the classes in both contexts correctly~\citep{kundu2020class}. 

For DIOD, a recent competition~\citep{ssladICCV} showed through their winning solutions that general strategies that account mainly for classification might suffice (e.g., simple random replay, using larger networks)~\citep{li2022technical, acharya20212nd, zhaitechnical} even in challenging scenarios. For that, we advise the reader to analyze the general findings and discussions present in related surveys and review papers~\citep{parisi2019continual, hadsell2020embracing, delange2021continual}. Contrastively, the CIOD paradigm needs a more specific treatment due to its inherent challenges and complexity. 

The task of incrementally adding classes to a trained detector is considered of substantial importance for several applications that deal with memory and computational constraints~\citep{shaheen2021continual}. The main issue that makes detection a more difficult task than only classification for class-incremental scenarios is that the same image can have several instances of different objects that are unknown apriori. Since these objects are not identified, the network learns to treat their visual cues as background instances. Later, when images of the unknown instances present before are shown as a new class, the model tends to either not converge to a decent solution or only prioritize the learning of the new category. In other words, this label conflict favors the interference on the weights specific to each task within the network.

\begin{figure}[!ht]
    \centering
        \begin{subfigure}[b]{0.49\columnwidth}
            \centering
            \includegraphics[width=\columnwidth]{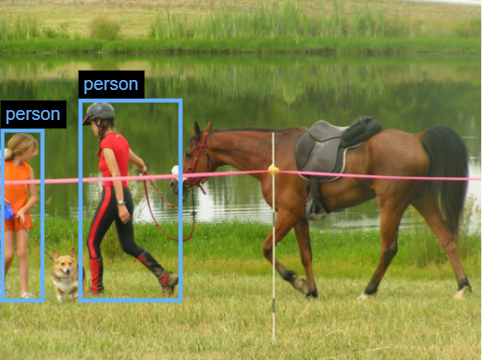}
            \caption{Learning $t_1$}
            \label{fig:incremental_classes_a}
        \end{subfigure}
    \hfill
        \begin{subfigure}[b]{0.49\columnwidth}
            \centering
            \includegraphics[width=\columnwidth]{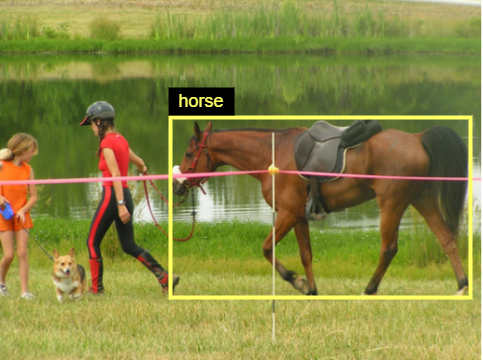}
            \caption{Learning $t_2$}
            \label{fig:incremental_classes_b}
        \end{subfigure}
    \hfill \\
        \begin{subfigure}[b]{0.49\columnwidth}
            \centering
            \includegraphics[width=\columnwidth]{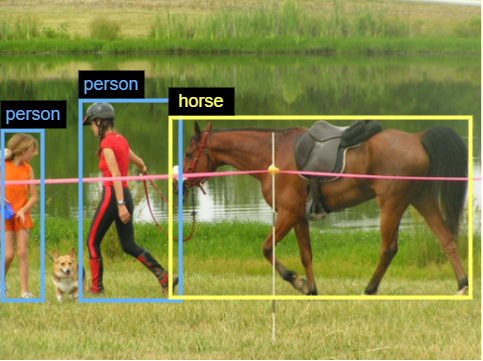}
            \caption{Incrementally learning $t_2$ after $t_1$}
            \label{fig:incremental_classes_c}
        \end{subfigure}
    \hfill
        \begin{subfigure}[b]{0.49\columnwidth}
            \centering
            \includegraphics[width=\columnwidth]{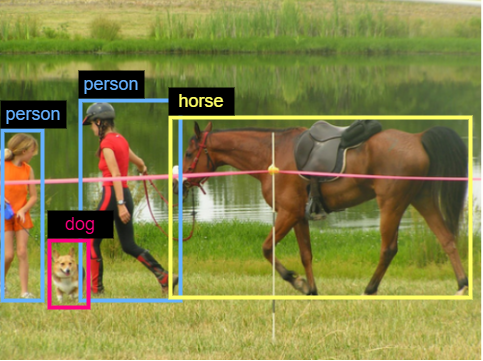}
            \caption{Incrementally learning a $t_3$ after $[t_1, t_2]$.}
            \label{fig:incremental_classes_d}
        \end{subfigure}
    \caption{Examples of learning separately some task $t_1$ and $t_2$; and incrementally learning $[t_1, t_2, t_3]$.}
    \label{fig:incremental_classes}
\end{figure}

Figure \ref{fig:incremental_classes} exemplifies the process of incrementally learning some classes, after a previous one was already learned. Figures \ref{fig:incremental_classes_a} and \ref{fig:incremental_classes_b} shows an example of two classes being learned separately, whilst Figure \ref{fig:incremental_classes_c} shows the new class for task $t_2$ being learned after $t_1$, at last Figure \ref{fig:incremental_classes_d} shows a third class added to the model on the task $t_3$. To exemplify why CIOD is considered a harder task than classification, the class ``person'' for the first learning task represented on Figure \ref{fig:incremental_classes_a} is considered as background on the second and third tasks (Figure \ref{fig:incremental_classes_b}). This naturally results in a label conflict that might induce catastrophic forgetting and harm the final detection performance. 

Although still in its first steps, the CIOD field has a more established corpus of strategies and some of them can also be applied within the domain-incremental option~\citep{li2022technical, kundu2020class}. The first proposed strategy for CIOD dates back to 2017 in the seminal paper written by \citet{shmelkov2017incremental}. Since then, several methods have been presented with the goal of tackling forgetting while making DNNs localize and recognize classes incrementally. For a more concise way of analyzing all the recent contributions to this field, we performed a systematic review of all the papers that included evaluations within the scope of continual object detection for class-incremental scenarios.

\subsection{Considerations about the Literature Review}

\begin{figure}[h]
    \centering
    \includegraphics[width=\columnwidth]{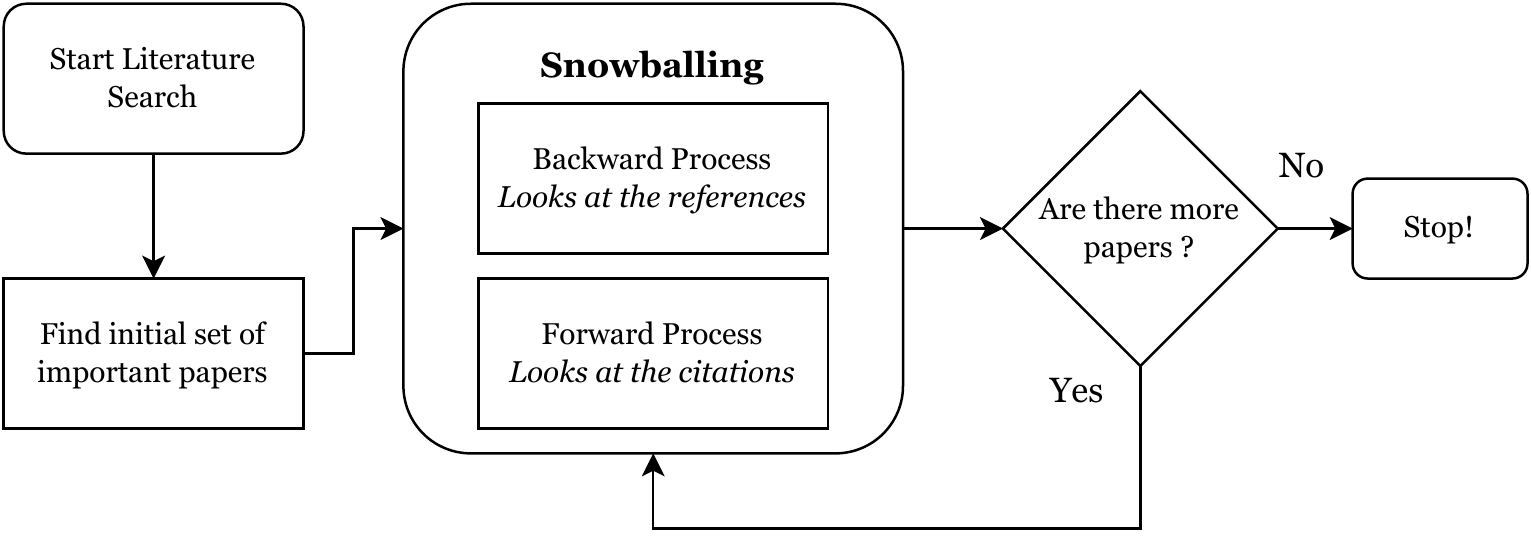}
    \caption{The adopted snowballing review process.}
    \label{fig:snowball}
\end{figure}

For gathering the most influential work related to the CIOD field, we took advantage of the fact that the initial paper of ~\citet{shmelkov2017incremental} presented a solid baseline for the problem, which indirectly guided the field to always make comparisons to it. In this way, we chose to perform a snowballing literature review followed by the guidelines described on ~\citet{wohlin2014guidelines}. In this review technique, a paper (or a set of papers) has its citations and references explored in a forward and backward iterative process in order to find all works that deal with the topic of interest. A general description of the review pipeline is described in Figure~\ref{fig:snowball}.

Since the research field is reasonably new, some relevant work will certainly be placed first on arXiv as pre-prints. Because of that, we decided to use the Google Scholar database for checking the citations and references since they aggregate all the results from pre-print sources (e.g., arXiv and bioRxiv) to several popular scientific databases such as IEEE Xplore, ACM Digital Library, Scopus, and Science Direct.

\subsubsection{Research Questions}

With this review, we aimed to answer the following research questions:

\begin{description}
    \item[RQ1:] What are the main proposed strategies for CIOD ?
    \item[RQ2:] What are the main benchmarks ?
    \item[RQ3:] What are the main metrics ?
    \item[RQ4:] What is the current state-of-the-art with respect to performance ?
\end{description}

\subsubsection{Inclusion and Exclusion criteria}

For starting the forward and backward process inherent to the snowballing technique, we considered the following inclusion criteria:

\begin{enumerate}[label=\checkmark]
    \item Papers that cited \citet{shmelkov2017incremental} or appeared in its reference list.
\end{enumerate}

Then, we iteratively checked all the papers that made citations (up to March 2022) or appeared in the reference list of the first pool of gathered papers and proceeded in a loop until no more studies could be considered. At the same time, for selecting the works that mattered to this proposal from this large set, we established the following exclusion criteria:

\begin{enumerate}[label=$\times$]
    \item Paper was not written in English.
    \item Paper did not propose a technique, benchmark or metric related to the CIOD paradigm.
    \item Paper did not go under the peer-review process or, if published as pre-print online, did not have citations.
\end{enumerate}

As stated above, we adopted the requirement for citations as a quality measure only for the works published as pre-prints online. This strategy was adopted considering that the CIOD field is recent (i.e., many papers will be placed as pre-prints before being published), and we value public acceptance as a way to evaluate a paper`s integrity. After analyzing all related work, 26 research papers followed the criteria and provided answers to the aspects indicated by the aforementioned research questions.

\subsection{Literature Review Results}
\label{rw-techniques}

In this subsection we proceed with the discussion of the review results and the formulation of answers to each research question.

\subsubsection{RQ1: What are the main proposed strategies for CIOD?}
\label{sec:RQ1}
For dissecting the contributions of each paper, we evaluated the selected works on the choice of strategy to mitigate forgetting, used architecture and backbone, benchmarks and evaluation methods. The results are presented in Table~\ref{tab-CIOD-papers} with some colored cells to aid in the analysis. 

\begin{table*}
\centering
\caption{Class-Incremental Object Detection main papers}\label{tab-CIOD-papers}
\scalebox{0.9}{
\begin{tabular}{llllll} 
\hline
References                                                             & Strategy                                                         & Benchmark                                                                             & Backbone                                    & Object Detector                                     & Evaluation                                                \\ 
\hline
\begin{tabular}[c]{@{}l@{}}\citet{shmelkov2017incremental} (ILOD)\end{tabular}     & \colorbox{yellow}{Knowledge Distillation}                                                               & \begin{tabular}[c]{@{}l@{}}VOC 2007\\COCO 2014\end{tabular}                                               & ResNet-50                                                       & Fast-RCNN                                                                & \begin{tabular}[c]{@{}l@{}}Multiple Classes\\Sequential Classes\end{tabular}  \\
\citet{li2018incremental} (MMN)                                                      & \colorbox{cyan}{Parameter Isolation}                                                                  & VOC 2007                                                                                                  & VGG-16                                                          & SSD-300                                                                  & \begin{tabular}[c]{@{}l@{}}Multiple Classes\\Sequential Classes\end{tabular}  \\
\citet{guan2018learn}                                                                & \colorbox{lime}{Pseudo-Labels}                                                                        & \begin{tabular}[c]{@{}l@{}}VOC 2007\\TSD-MAX\end{tabular}                                                 & Darknet-19                                                      & Yolo-V2                                                                  & Multiple Classes                                                              \\
\citet{hao2019end} (CIFRCN)                                                          & \colorbox{yellow}{Knowledge Distillation}                                                               & \begin{tabular}[c]{@{}l@{}}VOC 2007\\COCO 2014\end{tabular}                                               & ResNet-101                                                      & \begin{tabular}[c]{@{}l@{}}Faster-RCNN + \\Nearest Neighbor\end{tabular} & \begin{tabular}[c]{@{}l@{}}Multiple Classes\\Sequential Classes\end{tabular}                                            \\
\citet{chen2019new}                                                                  & \colorbox{yellow}{Knowledge Distillation}                                                               & VOC 2007                                                                                                  & ResNet                                                          & Faster-RCNN                                                              & \begin{tabular}[c]{@{}l@{}}Multiple Classes\\Sequential Classes\end{tabular}  \\
\citet{li2019rilod} (RILOD)                                                          & \begin{tabular}[c]{@{}l@{}}\colorbox{yellow}{Knowledge Distillation}\\\colorbox{red}{External Data}\end{tabular}        & \begin{tabular}[c]{@{}l@{}}VOC 2007\\iKitchen\end{tabular}                                                & ResNet-50                                                       & RetinaNet                                                                & \begin{tabular}[c]{@{}l@{}}Multiple Classes\\Sequential Classes\end{tabular}  \\
\citet{hao2019take} (FCIOD)                                                          & \begin{tabular}[c]{@{}l@{}}\colorbox{yellow}{Knowledge Distillation}\\\colorbox{magenta}{Replay}\end{tabular}               & TGFS                                                                                                      & ResNet-101                                                       & Faster-RCNN                                                              & Multiple Classes                                                              \\
\citet{liu2020incdet} (IncDet)                                                       & \begin{tabular}[c]{@{}l@{}}\colorbox{lime}{Pseudo-Labels} \\\colorbox{brown}{EWC}\end{tabular}               & \begin{tabular}[c]{@{}l@{}}VOC 2007\\COCO 2014\end{tabular}                                               & ResNet-50                                                       & \begin{tabular}[c]{@{}l@{}}Fast-RCNN\\Faster-RCNN\end{tabular}           & \begin{tabular}[c]{@{}l@{}}Multiple Classes\\Sequential Classes\end{tabular}  \\
\begin{tabular}[c]{@{}l@{}}\citet{acharya2020rodeo} (RODEO)\end{tabular}           & \colorbox{magenta}{Replay}                                                                               & \begin{tabular}[c]{@{}l@{}}VOC 2007\\COCO 2014\end{tabular}                                               & ResNet-50                                                       & Fast-RCNN                                                                & Sequential Classes                                                            \\
\begin{tabular}[c]{@{}l@{}}\citet{peng2020faster} (Faster ILOD)\end{tabular}       & \colorbox{yellow}{Knowledge Distillation}                                                               & \begin{tabular}[c]{@{}l@{}}VOC 2007\\COCO 2014\end{tabular}                                               & ResNet-50                                                       & Faster-RCNN                                                              & \begin{tabular}[c]{@{}l@{}}Multiple Classes\\Sequential Classes\end{tabular}  \\
\citet{zhou2020lifelong}                                                             & \colorbox{yellow}{Knowledge Distillation}                                                               & \begin{tabular}[c]{@{}l@{}}VOC 2007\\COCO 2014\end{tabular}                                               & ResNet-50                                                       & Faster-RCNN                                                              & \begin{tabular}[c]{@{}l@{}}Multiple Classes\\Sequential Classes\end{tabular}  \\
\citet{zhang2020class} (DMC)                                                         & \begin{tabular}[c]{@{}l@{}}\colorbox{yellow}{Knowledge Distillation}\\\colorbox{red}{External Data}\end{tabular}        & VOC 2007                                                                                                  & \begin{tabular}[c]{@{}l@{}}ResNet-50 / \\ResNet-34\end{tabular} & RetinaNet                                                                & Multiple Classes                                                              \\
\citet{liu2020multi} (AFD)                                                           & \begin{tabular}[c]{@{}l@{}}\colorbox{yellow}{Knowledge Distillation}\\\colorbox{magenta}{Replay}\end{tabular}               & \begin{tabular}[c]{@{}l@{}}KITTI / Kitchen \\VOC 2007 \\COCO 2014 \\Comic / Watercolor\end{tabular} & SE-ResNet-50                                                    & Faster-RCNN                                                              & Multiple Classes                                                              \\
\citet{yang2020two}                                                                  & \begin{tabular}[c]{@{}l@{}}\colorbox{lime}{Pseudo-Labels}\\\colorbox{yellow}{Knowledge Distillation}\end{tabular}        & \begin{tabular}[c]{@{}l@{}}VOC 2007\\COCO 2014\end{tabular}                                               & ResNet-50                                                       & Faster-RCNN                                                              & \begin{tabular}[c]{@{}l@{}}Multiple Classes\\Sequential Classes\end{tabular}  \\
\citet{shieh2020continual}                                                           & \colorbox{magenta}{Replay}                                                                               & \begin{tabular}[c]{@{}l@{}}VOC 2007\\ITRI-DriveNet-60\end{tabular}                                        & Darknet-53                                                      & Yolo-V3                                                                  & Multiple Classes                                                              \\
\begin{tabular}[c]{@{}l@{}}\citet{ramakrishnan2020relationship} (RKT)\end{tabular} & \colorbox{yellow}{Knowledge Distillation}                                                               & \begin{tabular}[c]{@{}l@{}}VOC 2007\\VOC 2012\\KITTI\end{tabular}                                         & ResNet                                                          & Fast-RCNN                                                                & \begin{tabular}[c]{@{}l@{}}Multiple Classes\\Sequential Classes\end{tabular}                                \\
\citet{chen2020incremental}                                                          & \colorbox{yellow}{Knowledge Distillation}  & DOTA / DIOR                                                                                               & \begin{tabular}[c]{@{}l@{}}Custom with \\FPN\end{tabular}       & \begin{tabular}[c]{@{}l@{}}Custom with \\two stages\end{tabular}         & Multiple Classes                                                              \\
\citet{peng2021sid} (SID)                                                            & \colorbox{yellow}{Knowledge Distillation}                                                               & \begin{tabular}[c]{@{}l@{}}VOC 2007\\COCO 2014\end{tabular}                                               & ResNet-50                                                       & \begin{tabular}[c]{@{}l@{}}CenterNet\\FCOS~ ~\end{tabular}               & \begin{tabular}[c]{@{}l@{}}Multiple Classes\\Sequential Classes\end{tabular}  \\
\citet{joseph2021towards} (ORE)                                                     & \begin{tabular}[c]{@{}l@{}}\colorbox{lime}{Pseudo-Labels} \\\colorbox{magenta}{Replay}\end{tabular}            & VOC 2007                                                                                                  & ResNet-50                                                       & \begin{tabular}[c]{@{}l@{}}Faster-RCNN + \\Nearest Neighbor\end{tabular} & Multiple Classes                                                              \\
\citet{yang2021multi}~                                                               & \colorbox{yellow}{Knowledge Distillation}                                                               & \begin{tabular}[c]{@{}l@{}}VOC 2007\\COCO 2014\end{tabular}                                               & ResNet-50                                                       & Faster-RCNN                                                              & \begin{tabular}[c]{@{}l@{}}Multiple Classes\\Sequential Classes\end{tabular}  \\
\citet{yang2021objects}                                                              & \colorbox{magenta}{Replay}                                                                               & VOC 2007                                                                                                  & ResNet-50                                                       & Faster-RCNN                                                              & Multiple Classes                                                              \\
\citet{kj2021incremental} (Meta-ILOD)                                                & \begin{tabular}[c]{@{}l@{}}\colorbox{yellow}{Knowledge Distillation}\\\colorbox{magenta}{Replay} \colorbox{teal}{Meta-Learning}\end{tabular} & \begin{tabular}[c]{@{}l@{}}VOC 2007\\COCO 2014\end{tabular}                                               & ResNet-50                                                       & Faster-RCNN                                                              & \begin{tabular}[c]{@{}l@{}}Multiple Classes\\Sequential Classes\end{tabular}  \\
\citet{ul2021incremental}                                                            & \colorbox{yellow}{Knowledge Distillation}                                                               & VOC 2007                                                                                                  & Darknet-53                                                      & Yolo-V3                                                                  & Sequential Classes                                                            \\
\citet{zhang2021incremental}                                                         & \colorbox{cyan}{Parameter Isolation}                                                                  & VOC 2007                                                                                                  & \begin{tabular}[c]{@{}l@{}}Darknet-53 +\\ResNet\end{tabular}    & Yolo-V3                                                                  & Sequential Classes                                                            \\
\citet{dong2021bridging}                                                             & \begin{tabular}[c]{@{}l@{}}\colorbox{yellow}{Knowledge Distillation}\\\colorbox{red}{External Data}\end{tabular}                                                               & \begin{tabular}[c]{@{}l@{}}VOC 2007\\COCO 2014\end{tabular}                                               & ResNet-50                                                       & Faster-RCNN                                                              & \begin{tabular}[c]{@{}l@{}}Multiple Classes\\Sequential Classes\end{tabular}  \\
\citet{wang2021wanderlust}                                                           & \multicolumn{1}{c}{-}                                                                & OAK                                                                                                       & ResNet-50                                                       & Faster-RCNN                                                              & Sequential Classes                                                            \\
\hline
\end{tabular}}
\end{table*}

\vspace{0.2cm}
\noindent
\colorbox{yellow}{\textbf{Knowledge Distillation}}
\vspace{0.2cm}

As mentioned previously, \citet{shmelkov2017incremental} introduced the first work to deal with the CIOD problem through the use of ``vanilla'' knowledge distillation. The authors adapted the Fast-RCNN architecture to learn incrementally by using a copy of the network trained on the base classes as the teacher and another as the student. The teacher has its weights frozen and the student has to not only detect the newly introduced categories but also repeat the distribution of responses of the frozen teacher. This behavior is achieved by using an additional regularization loss based on the bounding box predictions and logits produced by both networks, inspired by the work of ~\citet{li2017learning}. Since they were the first to propose a strategy for this problem, most consecutive papers built solutions on top of their initial regularization approach and compared them to it. 

\citet{hao2019end} adapted the Faster-RCNN architecture to the CIOD context with the expansion of the RPN to consider the new class as foreground. They evaluated the classification results using a fully connected network and a nearest prototype classifier. Additionally, they artificially avoided the possibility of background label conflict between old and new data by excluding images that contained objects from multiple class groups, which is unreal for a real-world setting. In a similar strategy,~\citet{chen2020incremental} expanded the RPN for dealing with new classes and used knowledge distillation on the outputs of a teacher network to allow the model to detect remote sensing objects incrementally with minimum forgetting using the specific domain datasets proposed by \citet{li2020object} and \citet{xia2018dota}.~\citet{zhou2020lifelong} applied distillation on the detection heads and RPN outputs along with a supplementary sampling strategy to select proposals that tend to be from the foreground classes.~\citet{ramakrishnan2020relationship} hypothesized that the relationship between region proposals and the ground truth annotations encoded the detector`s knowledge. In this way, the authors introduced a strategy to select proposals based on their relation and applied distillation on the filtered samples.~\citet{ul2021incremental} evaluated distilling knowledge only on the logits for the YOLO-V3 architecture in a setting with two classes and showed better results than other CL strategies.

Beyond the basic distillation of the detector outputs, several methods proposed additionally to distill intermediate features of the base model. \citet{chen2019new} presented the first work that made use of this type of distillation through what they named a ``hint loss'', but they provided limited results of their approach.~\citet{peng2020faster} made use of the Faster-RCNN and introduced an additional adaptive distillation step on the features and RPN outputs. They additionally investigated the negative impact that having old class objects within the new class images has on the performance of the RPN and concluded that it was not that significant, which explains why Faster-RCNN networks generalize better than solutions with external proposals.~\citet{peng2021sid} presented the use of distillation not only on intermediate features, but also on the relations (distances) between features of different samples for anchor-free object detectors.~\citet{yang2021multi} proposed the preservation of channel-wise, point-wise, and instance-wise correlations between some feature maps of the teacher and student networks in order to maintain the performance on the old classes while optimizing for the new ones.

\vspace{0.2cm}
\noindent
\colorbox{magenta}{\textbf{Replay}}
\vspace{0.2cm}

\citet{hao2019take} employed the use of a small buffer of samples along with logits distillation to perform better than its competitors in the incremental learning of common objects from vending machines.~\citet{shieh2020continual} proposed the use of experience replay with different buffer sizes and the YOLO-V3 architecture for the problem of adding multiple classes at once to an object detector. They evaluated their approach in a common benchmark and on a private autonomous driving dataset.~\citet{acharya2020rodeo} suggested the use of product quantization to compress feature maps without losing their fine-grained resolution, which allowed for keeping a low-memory profile while performing well on some incremental benchmarks.~\citet{liu2020multi} presented the use of an adaptive exemplar sampling for selecting replay instances and proposed different ways of applying the attention mechanism within the feature distillation procedure as a strategy to hinder forgetting. They evaluated their approach on various benchmarks and diverse scenarios in which the incremental data did not share the same domain as the base classes.~\citet{yang2021objects} proposed the use of a pre-trained language model to constrain the topology of the feature space within the model and capture the nuances of semantic relations associated with each class name. Their solution was meant to be used for open-world object detection. Still, it can also deal with incremental detection by using a replay buffer with prototypes for each class to prevent forgetting old categories. 

\vspace{0.2cm}
\noindent
\colorbox{cyan}{\textbf{Parameter Isolation}}
\vspace{0.2cm}

~\citet{li2018incremental} introduced a simple strategy for dealing with forgetting based on ``mining'' important network parameters and freezing them. For each task, they sorted the weight parameters by their magnitude and stored their values and positions in a memory buffer so that when training for the next task, the parameters would be reset to their original values.~\citet{zhang2021incremental} proposed a compositional architecture based on the mixture of compact expert detectors. They trained a YOLO-V3 network using a sparse mechanism for each detection task and then applied the pruning technique suggested by~\citet{liu2017learning} to eliminate unimportant channels and residual blocks. For selecting which expert to forward the inputs, they used a ResNet-50 classifier as the ``oracle''. Their strategy presented interesting results since the final model was able to keep a low memory footprint and no forgetting of old classes. Yet, their system was evaluated in a limited scenario with only three incremental tasks, making it difficult to compare to other techniques.

\vspace{0.2cm}
\noindent
\colorbox{lime}{\textbf{Pseudo-Labels}}
\vspace{0.2cm}

\citet{guan2018learn} showed that when the base classes instances are also present in the images of the incremental categories, self-labeling using the own model could be a good enough strategy for dealing with forgetting.~\citet{liu2020incdet} identified that pseudo-labels are an essential step when one wants to regularize the weight of a network with EWC. Moreover, they also introduced a novel Huber regularization loss for constraining the gradients of each parameter based on their relevance to the old classes.~\citet{yang2020two} presented the use of pseudo-labels on the new classes images along with the application of general feature and output distillation and the learning of a residual model to compensate for the discrepancies between the teacher and student networks.~\citet{joseph2021towards} suggested the application of self-labeling to identify potential unknown objects on an image for open-world object detection. To prevent forgetting, they save a replay buffer with class prototypes and apply contrastive clustering in the feature space so that new classes can be added sequentially.

\vspace{0.2cm}
\noindent
\colorbox{red}{\textbf{External Data}}
\vspace{0.2cm}

~\citet{li2019rilod} used a one-stage detector (RetinaNet) and not only distilled the knowledge of outputs and intermediate features but also idealized a way to automatically collect and annotate new data from search engines such as the Google Image Search tool to be used during the incremental training and testing schemes for improved performance. ~\citet{zhang2020class} proposed the independent training of one-stage networks on the base and new classes and the further transfer of their specific knowledge to a new separate network via knowledge distillation using an external unlabeled dataset.~\citet{dong2021bridging} explored the scenario of non co-occurrence of old classes in new classes images. They proposed a blind sampling strategy to select samples from large labeled in-the-wild datasets (e.g., COCO). To prevent forgetting, they designed a distillation strategy based on the remodeled output of the detection head, RoI masks on the image-level, and heatmaps on the instance-level.

\vspace{0.2cm}
\noindent
\vspace{0.2cm}
\colorbox{teal}{\textbf{Meta-Learning}}

~\citet{kj2021incremental} produced a hybrid strategy that relied on knowledge distillation, replay, and meta-learning to avoid forgetting. Along with the use of knowledge distillation on the outputs and backbone features, they used the gradient conditioning technique proposed by \citet{flennerhag2019meta} to regularize the weight updates on some layers of the detector RoI head. This technique granted the ability of fast adaptation by fine-tuning with data from new classes and a few samples of old ones from a replay buffer.

\subsubsection{RQ2: What are the main benchmarks?}\label{RQ2}

\begin{figure}[h]
\centering
\includegraphics[width=\linewidth]{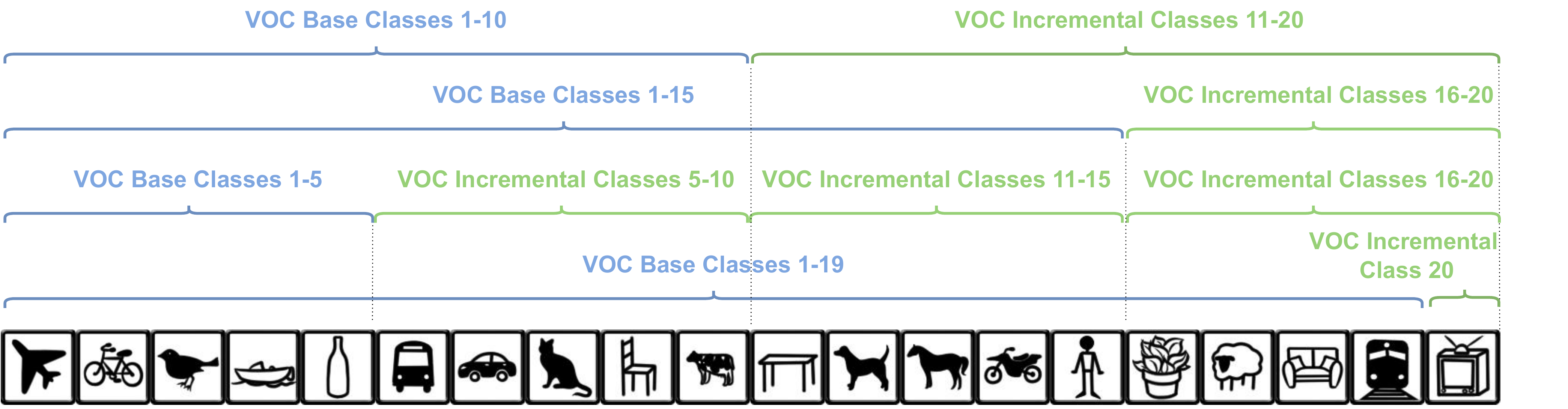}
\caption{Description of some of the adopted incremental scenarios for the Pascal VOC 2007 dataset.}
\label{fig-VOC-IOD}
\end{figure}

Every benchmark designed for general object detection can be adjusted for the class-incremental paradigm. This can be achieved by allowing the model only to see the instances from the classes of interest, making sure to omit the annotations related to the categories that are not part of the current experience. Most of the strategies presented in Table~\ref{tab-CIOD-papers} were compared using the adaptation of the traditional VOC and COCO benchmarks with some caveats to make them incremental by introducing classes sequentially in single units or pre-defined groups. For the incremental version of the VOC dataset, classes were alphabetically ordered and generally split into four different scenarios, as illustrated by Figure~\ref{fig-VOC-IOD}.

\begin{figure}[!h]
\centering
\includegraphics[width=\linewidth]{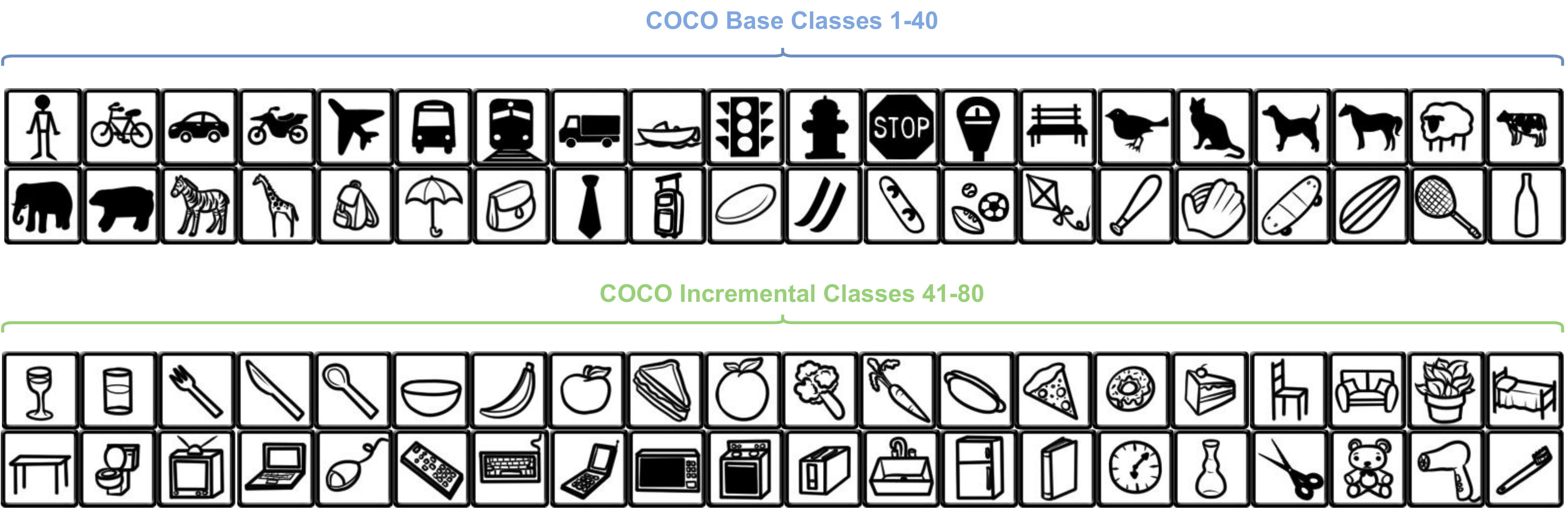}
\caption{Description of an incremental scenario with MS COCO 2014 dataset.}
\label{fig-COCO-IOD}
\end{figure}

For creating the incremental scheme for the COCO dataset, classes are ordered following the ID of the original labels and split in half to create a unique scenario with 40 classes for training the base model and 40 to be added sequentially as shown in Figure~\ref{fig-COCO-IOD}.

Although not well explored, there were a few datasets designed for evaluating CIOD solutions.~\citet{hao2019take} introduced a large-scale dataset of vending machine products with 38k images and 24 possible categories called Take Goods from Shelves (TGFS). The benchmark also has three coarse classes that cover the categories and was meant to instigate class-incremental detection solutions to retail problems.~\citet{wang2021wanderlust} proposed an egocentric video dataset that focused on capturing objects and scenes present in the daily life of a university student. The benchmark is called Objects Around Krishna (OAK) and was meant to be used for online continual object detection tasks.

\subsubsection{RQ3: What are the main metrics?}\label{RQ3}

The evaluation for CIOD has followed the same structure of traditional object detection with the use of $mAP@.5$ for nearly all benchmarks and $mAP@0.5-0.95$ for COCO like datasets. However, some researchers noticed that directly comparing the $mAP$ performance of techniques on the same benchmark would not assess their real efficiency since changes in the training regime, and even framework could cause the same method to present different results. To comply with that, the difference and the ratio against the upper-bound (i.e., joint-training with all classes at once) have been commonly used for comparisons~\citep{liu2020incdet, acharya2020rodeo} since they represent how the performance would be in case data could be fully accumulated and create a common ground between techniques (i.e., how far we are from the ideal response). Yet, the gap against the joint-training is only meaningful when both methods are implemented within the same training regime and framework since only in this situation it is possible to ascertain which single components really contributed to narrowing the gap. Most researchers do not consider this setting and pick up results from different papers to compare against their joint-training outcomes, which does not give more information than checking their single $mAP$ results.

Beyond that,~\citet{chen2019new} proposed the use of a $F_{map}$ metric, inspired by the $F_{1}-score$, in which they calculate the harmonic mean between the $mAP$ values of old and new classes as described by Equation~\ref{eq-fmap}.

\begin{equation}
    F_{map}= \frac{2\ \ mAP_{old}\ \ mAP_{new}}{mAP_{old}+ mAP_{new}}
    \label{eq-fmap}
\end{equation}

~\citet{yang2021multi} introduced a metric called Stability-Plasticity-mAP $SPmAP$ that considers how much the incremental learning process affects the average stability and plasticity of a detector. Their metric takes into consideration the mean differences of the incremental model against the upper-bound for the old and new classes as shown by Equation~\ref{eq-spmap}.

\begin{equation}
    \label{eq-spmap}
    SPmAP = \frac{\frac{Stability + Plasticity}{2} + mAP_{dif}}{2}
\end{equation}

\begin{equation}
    Stability= \frac{1}{N_{old\_classes}}\sum_{i=1}^{N_{old\_classes}}(mAP_{joint,i}-mAP_{inc,i}) \notag
\end{equation}

\begin{equation}
    Plasticity= \frac{1}{N_{new\_classes}}\sum_{i=N_{old\_classes}+1}^{N_{all\_classes}}(mAP_{joint,i}-mAP_{inc,i}) \notag
\end{equation}

\begin{equation}
    mAP_{dif}= \frac{1}{N_{all\_classes}}\sum_{i=1}^{N_{all\_classes}}(mAP_{joint,i}-mAP_{inc,i}) \notag
\end{equation}

We also believe that CIOD models can only be compared when the join-training results of their architecture are available. However, only looking at the discrepancy between the incremental and joint-training models does not lead to the understanding of which specific aspects of the strategy are failing. The aforementioned metrics are helpful, but they lack the specificity for identifying where the incremental model should pay attention. To circumvent that, we propose two separate metrics that compare and scale the final incremental $mAP$ values for each class against the joint-training separately for the old and new categories. These metrics are defined as the rate of stability (RSD) and plasticity (RPD) deficits as described in Equations~\ref{eq-rate-stab} and ~\ref{eq-rate-plast}.


\begin{equation}
    \label{eq-rate-stab}
    \text{RSD} = \frac{1}{N_{old\_classes}} \sum_{i=1}^{N_{old\_classes}}\frac{mAP_{joint,i} - mAP_{inc,i}}{mAP_{joint,i}} \ * 100
\end{equation}


\begin{equation}
    \label{eq-rate-plast}
    \text{RPD} = \frac{1}{N_{new\_classes}} \sum_{i=N_{old\_classes}+1}^{N_{new\_classes}}\frac{mAP_{joint,i} - mAP_{inc,i}}{mAP_{joint,i}} \ * 100
\end{equation}

Our metrics allow the direct interpretation of how much an incremental model compares to the upper-bound in remembering old classes and learning the new ones (e.g., the model has a 10\% worse performance for recognizing previous classes, but only a 2\% deficit for learning new categories when compared to the upper-bound). Therefore, for this context, a CIOD strategy should aim not only to reach a decent final $mAP$ value and high upper-bound ratio but also to keep low and balanced stability and plasticity deficits. Additionally, the ratios can assume negative values, indicating that the incremental model has performed better than joint-training for some classes and reinforcing the relationship with standard CL metrics such as BWT. We applied the rate of plasticity and stability deficits along with the upper-bound difference during the performance analysis of the following section. 

\subsubsection{RQ4: What is the current state-of-the-art with respect to performance?}\label{sec-REQ4-perform}

For performing an investigation on which strategies have worked better for CIOD, we need to find common ground among them. Yet, there are no standards for frameworks, architecture backbones, and training regimes regarding paper reimplementations. Some papers tried to replicate the number of iterations, learning rates, and procedures used by \citet{shmelkov2017incremental}, but there is a clear difference in the obtained results that can be seen mainly for the joint training cases that used the same architecture~\citep{peng2020faster, hao2019end}. In this way, it is difficult to state that some results are better than others because of the proposed policies and not due to the better selection of hyperparameters, which has been shown previously to highly influence generalization in the CL setting~\citep{mirzadeh2020understanding}. Therefore, using a consistent evaluation procedure is essential for identifying the most promising directions in the field.

Tables~\ref{voc-mul},~\ref{voc-seq} and ~\ref{coco-mul}, present the results of each paper that was evaluated on the PASCAL VOC 2007 and MS COCO 2014 following the main benchmarks described in Section~\ref{RQ2} for when multiple and singles classes are added sequentially. The metrics proposed in Section~\ref{RQ3} are used for evaluating the real impact of each strategy according to their upper-bound. Because our metrics also need access to the $mAP$ of each class for both the incremental and joint-training models, some previously discussed works have a $\dagger$ symbol that indicates that the paper only provided the mean $mAP$ value for the old and new classes in groups for each setting. 

In Table~\ref{voc-mul}, by looking at the final $mAP$ and the values of the upper-bound ratios for all incremental scenarios, it is possible to conclude that for the VOC benchmark, as more classes are added at once, the more complex the task becomes for the detectors. Strategies based on pseudo-labels and replay demonstrated consistent results. In contrast, pure knowledge distillation based techniques struggled more and had an average plasticity deficit of more than 10\%, which might be an indication that this type of regularization needs to be adjusted carefully for not harming the learning of new categories. Nevertheless, for the settings with 5 and 10 classes added at once, although the initial baseline from ~\citet{shmelkov2017incremental} presented a final $mAP$ often below its competitors, it also offered a good balance of stability and plasticity according to its upper-bound, which contributes to why this technique is still relevant for comparisons.

\begin{table}
\centering
\caption{VOC 2007 results for one or multiple classes added at once}
\label{voc-mul}
\scalebox{0.7}{
\begin{tabular}{llcccc}
\hline
                                      &                      & \multicolumn{4}{c}{\multirow{2}{*}{VOC 2007 Incremental (1-19 + 20)}}                                                                                                                                                                                      \\
                                      &                      & \multicolumn{4}{c}{}                                                                                                                                                                                                                                       \\ 
\hline
\multirow{2}{*}{Paper}                &                      & \multicolumn{1}{l}{\multirow{2}{*}{Final mAP}} & \multicolumn{1}{c}{\multirow{2}{*}{$\Omega_{all} \uparrow$}}  & \multicolumn{1}{c}{\multirow{2}{*}{RSD (\%) $\downarrow$}}          & \multicolumn{1}{c}{\multirow{2}{*}{RPD (\%) $\downarrow$}}            \\
                                      &                      & \multicolumn{1}{l}{}                           & \multicolumn{1}{l}{}                                        & \multicolumn{1}{l}{}                                                & \multicolumn{1}{l}{}                                                  \\ 
\hline
\begin{tabular}[c]{@{}l@{}}\citet{shmelkov2017incremental} (ILOD)\end{tabular}     &                      & 68.40     & 0.980                                                      & 1.90                                                                & 21.11                                                                 \\
\citet{li2018incremental} (MMN)                                                     &                      & 77.50     & 0.991                                                      & 1.09                                                                & -4.24                                                                 \\
\citet{li2019rilod} (RILOD)                                                          &                      & 65.00     & 0.870                                                      & 10.93                                                               & 48.67                                                                 \\
\begin{tabular}[c]{@{}l@{}}\citet{peng2020faster} (Faster ILOD)$\dagger$\end{tabular}       &                      & 68.56     & 0.972                                                      & 0.60                                                                & 44.27                                                                 \\
\citet{zhou2020lifelong}$\dagger$                                                            &                      & 69.60     & 0.991                                                      & -0.45                                                               & 24.82                                                                 \\
\citet{zhang2020class} (DMC)                                                        &                      & 70.80     & 0.948                                                      & 4.80                                                                & 12.33                                                                 \\
\citet{yang2020two}                                                                 &                      & 72.13     & 0.977                                                      & 0.93                                                                & 29.14                                                                 \\
\citet{shieh2020continual}                                                          &                      & 68.90     & 0.941                                                      & 3.41                                                                & 53.56                                                                 \\
\begin{tabular}[c]{@{}l@{}}\citet{ramakrishnan2020relationship} (RKT)\end{tabular} &                      & 67.20     & 0.984                                                      & 1.00                                                                & 14.29                                                                 \\
\citet{peng2021sid} (SID)$\dagger$                                                           &                      & 68.30     & 0.954                                                      & 4.61                                                                & 4.61                                                                  \\
\citet{joseph2021towards} (ORE)                                                    &                      & 68.89     & 0.977                                                      & 1.66                                                                & 14.51                                                                 \\
\citet{yang2021objects}$\dagger$                                                             &                      & 69.82     & 0.990                                                      & 0.41                                                                & 11.64                                                                 \\
\citet{yang2021multi}                                                               &                      & 69.70     & 0.973                                                      & 2.11                                                                & 12.17                                                                 \\
\citet{kj2021incremental} (Meta-ILOD)                                               &                      & 70.20     & 0.934                                                      & 5.82                                                                & 21.74                                                                 \\
\citet{dong2021bridging}$\dagger$                                                            &                      & 72.20     & 0.999                                                      & -1.38                                                               & 29.88                                                                 \\ 
\hline
                                      &                      & \multicolumn{4}{c}{\multirow{2}{*}{VOC 2007 Incremental (1-15 + 16-20)}}                                                                                                                                                                                      \\
                                      &                      & \multicolumn{4}{c}{}                                                                                                                                                                                                                                       \\ 
\hline
\multirow{2}{*}{Paper}                &                      & \multicolumn{1}{l}{\multirow{2}{*}{Final mAP}} & \multicolumn{1}{c}{\multirow{2}{*}{$\Omega_{all} \uparrow$}}  & \multicolumn{1}{c}{\multirow{2}{*}{RSD (\%) $\downarrow$}}          & \multicolumn{1}{c}{\multirow{2}{*}{RPD (\%) $\downarrow$}}            \\
                                      &                      & \multicolumn{1}{l}{}                           & \multicolumn{1}{l}{}                                        & \multicolumn{1}{l}{}                                                & \multicolumn{1}{l}{}                                                  \\ 
\hline
\begin{tabular}[c]{@{}l@{}}\citet{shmelkov2017incremental} (ILOD)\end{tabular}     &                      & 65.90     & 0.944                                                      & 3.60                                                                & 12.33                                                                 \\
\citet{liu2020incdet} (IncDet)$\dagger$                                                      &                      & 70.40     & 0.954                                                      & 0.44                                                                & 12.12                                                                 \\
\begin{tabular}[c]{@{}l@{}}\citet{peng2020faster} (Faster ILOD)$\dagger$\end{tabular}       &                      & 67.94     & 0.963                                                      & -3.60                                                               & 25.44                                                                 \\
\citet{yang2020two}                                                                 &                      & 69.71     & 0.944                                                      & 1.92                                                                & 17.75                                                                 \\
\citet{peng2021sid} (SID)$\dagger$                                                           &                      & 62.20     & 0.869                                                      & 13.13                                                               & 13.13                                                                 \\
\citet{joseph2021towards} (ORE)                                                    &                      & 68.51     & 0.972                                                      & 0.44                                                                & 10.71                                                                 \\
\citet{yang2021objects}$\dagger$                                                             &                      & 69.93     & 0.992                                                      & -3.55                                                               & 13.93                                                                 \\
\citet{yang2021multi}                                                               &                      & 66.50     & 0.929                                                      & 5.56                                                                & 11.92                                                                 \\
\citet{kj2021incremental} (Meta-ILOD)                                               &                      & 67.80     & 0.902                                                      & 6.58                                                                & 20.62                                                                 \\
\citet{dong2021bridging}$\dagger$                                                            &                      & 65.30     & 0.903                                                      & 2.49                                                                & 31.67                                                                 \\ 
\hline
                                      &                      & \multicolumn{4}{c}{\multirow{2}{*}{VOC 2007 Incremental (1-10 + 11-20)}}                                                                                                                                                                                      \\
                                      &                      & \multicolumn{4}{c}{}                                                                                                                                                                                                                                       \\ 
\hline
\multirow{2}{*}{Paper}                &                      & \multicolumn{1}{l}{\multirow{2}{*}{Final mAP}} & \multicolumn{1}{c}{\multirow{2}{*}{$\Omega_{all} \uparrow$}}  & \multicolumn{1}{c}{\multirow{2}{*}{RSD (\%) $\downarrow$}}          & \multicolumn{1}{c}{\multirow{2}{*}{RPD (\%) $\downarrow$}}            \\
                                      &                      & \multicolumn{1}{l}{}                           & \multicolumn{1}{l}{}                                        & \multicolumn{1}{l}{}                                                & \multicolumn{1}{l}{}                                                  \\ 
\hline
\begin{tabular}[c]{@{}l@{}}\citet{shmelkov2017incremental} (ILOD)
\end{tabular}     &                      & 63.10     & 0.904                                                      & 7.66                                                                & 11.42                                                                 \\
\citet{guan2018learn}                                                               &                      & 68.80     & 0.922                                                      & 11.06                                                               & 4.68                                                                  \\
\citet{chen2019new}$\dagger$                                                                 &                      & 33.50     & 0.474                                                      & 47.05                                                               & 69.02                                                                 \\
\citet{li2019rilod} (RILOD)                                                          &                      & 67.90     & 0.909                                                      & 10.42                                                               & 7.67                                                                  \\
\citet{liu2020incdet} (IncDet)$\dagger$                                                      &                      & 70.80     & 0.959                                                      & 4.52                                                                & 1.18                                                                  \\
\begin{tabular}[c]{@{}l@{}}\citet{peng2020faster} (Faster ILOD)$\dagger$\end{tabular}       &                      & 62.16     & 0.881                                                      & -4.79                                                               & 28.50                                                                 \\
\citet{zhou2020lifelong}$\dagger$                                                            &                      & 61.80     & 0.880                                                      & 9.16                                                                & 14.89                                                                 \\
\citet{zhang2020class} (DMC)                                                        &                      & 68.30     & 0.914                                                      & 7.63                                                                & 11.29                                                                 \\
\citet{yang2020two}                                                                 &                      & 66.21     & 0.897                                                      & 5.98                                                                & 14.74                                                                 \\
\citet{shieh2020continual}                                                          &                      & 65.50     & 0.895                                                      & 8.78                                                                & 14.04                                                                 \\
\begin{tabular}[c]{@{}l@{}}\citet{ramakrishnan2020relationship} (RKT)\end{tabular} &                      & 63.10     & 0.924                                                      & 1.25                                                                & 13.85                                                                 \\
\citet{peng2021sid} (SID)$\dagger$                                                           &                      & 59.80     & 0.835                                                      & 16.48                                                               & 16.48                                                                 \\
\citet{joseph2021towards} (ORE)                                                    &                      & 64.58     & 0.916                                                      & 14.76                                                               & 2.01                                                                  \\
\citet{yang2021objects}$\dagger$                                                             &                      & 64.96     & 0.921                                                      & 14.86                                                               & 0.89                                                                  \\
\citet{yang2021multi}                                                               &                      & 66.10     & 0.923                                                      & 7.29                                                                & 8.02                                                                  \\
\citet{kj2021incremental} (Meta-ILOD)                                               &                      & 66.30     & 0.882                                                      & 8.51                                                                & 15.07                                                                 \\
\citet{dong2021bridging}$\dagger$                                                            &                      & 59.90     & 0.828                                                      & 20.33                                                               & 13.97                                                                 \\
\hline
\end{tabular}}
\end{table}

Probably due to the increased complexity when working with several online updates, there were not many solutions to the sequential setting compared to its counterpart scenario, as shown in Table~\ref{voc-seq}. For when only five classes were being added sequentially, the parameter isolation strategy of~\citet{li2018incremental} demonstrated outstanding performance in the final $mAP$ and stability-plasticity metrics. Also, in their unique participation, the RODEO method from ~\citet{acharya2020rodeo} confirmed that replay is a suitable tool for dealing with consecutive one-class updates. Interestingly, the IncDet strategy from ~\citet{liu2020incdet} performed well on the setup of multiple groups being added sequentially but seemed to fail when learning single classes alone. This may be related to how strong the regularization penalty was adjusted to prevent the parameters from deviating much from the previously known distribution. In general, although having the same number of classes, the final $mAP$ was clearly lower in this setting when compared to adding the multiple categories at once. This corroborates that the ``tug-of-war'' on the parameters is happening actively in each new class network update.

\begin{table}[!htb]
\centering
\caption{VOC 2007 results for one or a group of classes added sequentially}
\label{voc-seq}
\scalebox{0.65}{
\begin{tabular}{llcccc} 
\hline
                                      &                      & \multicolumn{4}{c}{\multirow{2}{*}{VOC 2007 Incremental (1-15 + 16 + ... + 20)}}                                                                                                                                                                                      \\
                                      &                      & \multicolumn{4}{c}{}                                                                                                                                                                                                                                       \\ 
\hline
\multirow{2}{*}{Paper}                &                      & \multicolumn{1}{l}{\multirow{2}{*}{Final mAP}} & \multicolumn{1}{c}{\multirow{2}{*}{$\Omega_{all} \uparrow$}}  & \multicolumn{1}{c}{\multirow{2}{*}{RSD (\%) $\downarrow$}}          & \multicolumn{1}{c}{\multirow{2}{*}{RPD (\%) $\downarrow$}}            \\
                                      &                      & \multicolumn{1}{l}{}                           & \multicolumn{1}{l}{}                                        & \multicolumn{1}{l}{}                                                & \multicolumn{1}{l}{}                                                  \\ 

\hline
\begin{tabular}[c]{@{}l@{}}\citet{shmelkov2017incremental} (ILOD)\end{tabular}     &  & 62.40     & 0.894                                                      & 6.89                                                                & 22.56                                                                 \\
\citet{li2018incremental} (MMN)                                                      &  & 76.00     & 0.972                                                      & 2.21                                                                & 4.48                                                                  \\
\citet{liu2020incdet} (IncDet)$\dagger$                                                      &  & 67.60     & 0.916                                                      & 1.12                                                                & 35.87                                                                 \\
\citet{yang2020two}                                                                 &  & 59.62     & 0.807                                                      & 7.98                                                                & 56.45                                                                 \\
\citet{peng2021sid} (SID)$\dagger$                                                           &  & 48.90     & 0.683                                                      & 31.70                                                               & 31.70                                                                 \\
\citet{kj2021incremental} (Meta-ILOD)                                               &  & 65.70     & 0.874                                                      & 8.77                                                                & 25.08                                                                 \\ 
\hline
                                      &                      & \multicolumn{4}{c}{\multirow{2}{*}{VOC 2007 Incremental (1-10 + 11 + ... + 20)}}                                                                                                                                                                                      \\
                                      &                      & \multicolumn{4}{c}{}                                                                                                                                                                                                                                       \\ 
\hline
\multirow{2}{*}{Paper}                &                      & \multicolumn{1}{l}{\multirow{2}{*}{Final mAP}} & \multicolumn{1}{c}{\multirow{2}{*}{$\Omega_{all} \uparrow$}}  & \multicolumn{1}{c}{\multirow{2}{*}{RSD (\%) $\downarrow$}}          & \multicolumn{1}{c}{\multirow{2}{*}{RPD (\%) $\downarrow$}}            \\
                                      &                      & \multicolumn{1}{l}{}                           & \multicolumn{1}{l}{}                                        & \multicolumn{1}{l}{}                                                & \multicolumn{1}{l}{}                                                  \\ 

\hline
\citet{chen2019new}$\dagger$                                                                 &  & 33.50     & 0.474                                                      & 47.05                                                               & 69.02                                                                 \\
\begin{tabular}[c]{@{}l@{}}\citet{acharya2020rodeo} (RODEO)$\dagger$\end{tabular}           &  & 63.72     & 0.887                                                      & 13.78                                                               & 8.95                                                                  \\
\citet{zhou2020lifelong}$\dagger$                                                            &  & 46.20     & 0.658                                                      & 22.46                                                               & 45.82                                                                 \\ 
\hline
                                      &                      & \multicolumn{4}{c}{\multirow{2}{*}{VOC 2007 Incremental (1-5 + 6-10 + 11-15 + 16-20)}}                                                                                                                                                                                      \\
                                      &                      & \multicolumn{4}{c}{}                                                                                                                                                                                                                                       \\ 
\hline
\multirow{2}{*}{Paper}                &                      & \multicolumn{1}{l}{\multirow{2}{*}{Final mAP}} & \multicolumn{1}{c}{\multirow{2}{*}{$\Omega_{all} \uparrow$}}  & \multicolumn{1}{c}{\multirow{2}{*}{RSD (\%) $\downarrow$}}          & \multicolumn{1}{c}{\multirow{2}{*}{RPD (\%) $\downarrow$}}            \\
                                      &                      & \multicolumn{1}{l}{}                           & \multicolumn{1}{l}{}                                        & \multicolumn{1}{l}{}                                                & \multicolumn{1}{l}{}                                                  \\ 

\hline
\citet{hao2019end} (CIFRCN)                                                         &  & 48.50     & 0.694                                                     & 36.62                                                               & 12.09                                                                 \\
\citet{liu2020incdet} (IncDet)$\dagger$                                             &  & 62.60     & 0.848                                                     & 11.58                                                               & 21.67                                                                 \\
\citet{yang2020two}                                                                 &  & 49.05     & 0.664                                                     & 38.50                                                               & 3.00                                                                  \\
\begin{tabular}[c]{@{}l@{}}\citet{ramakrishnan2020relationship} (RKT)\end{tabular} &  & 52.90     & 0.775                                                     & 20.56                                                               & 29.23                                                                 \\
\citet{peng2021sid} (SID)$\dagger$                                                  &  & 36.20     & 0.506                                                     & 49.44                                                               & 49.44                                                                 \\
\citet{yang2021multi}                                                               &  & 27.66     & 0.386                                                     & 66.30                                                               & 44.77                                                                 \\
\hline
\end{tabular}}
\end{table}

Considering the results for the COCO incremental benchmark exhibited in Table~\ref{coco-mul}, techniques based on Faster-RCNN and feature distillation presented decent results. Even though the number of classes is higher than in the VOC benchmark, the upper-bound ratio shows that as the network is updated all at once, the forgetting condition is not as strong as in the sequential update example. Beyond that, methods based on self-labeling demonstrated results that justify their effectiveness for dealing with scenarios where a substantial number of classes was already introduced.

\begin{table}[!htb]
\centering
\caption{MS COCO 2014 results for multiple classes being added at once}
\label{coco-mul}
\scalebox{0.72}{
\begin{tabular}{llccc} 
\hline
                                      &  & \multicolumn{3}{l}{\multirow{2}{*}{MS COCO Incremental (1-40 + 41-80)}}                                                                                                                                    \\
                                      &  & \multicolumn{3}{l}{}                                                                                                                                                                                       \\ 
\hline
\multirow{2}{*}{Paper}                &  & \multicolumn{1}{l}{\multirow{2}{*}{mAP@.5}} & \multicolumn{1}{l}{\multirow{2}{*}{mAP@[.5, .95]}} & \multicolumn{1}{l}{\multirow{2}{*}{\begin{tabular}[c]{@{}l@{}}$\Omega_{all} \ [@.5]\uparrow$\end{tabular}}}  \\
                                      &  & \multicolumn{1}{l}{}                        & \multicolumn{1}{l}{}                               & \multicolumn{1}{l}{}                                                                                    \\ 
\hline
\begin{tabular}[c]{@{}l@{}}\citet{shmelkov2017incremental} (ILOD)\end{tabular} &  & 37.40  & 21.30         & 0.982                                                             \\
\citet{liu2020incdet} (IncDet)$\dagger$                                                  &  & 49.30  & 29.70         & 0.978                                                             \\
\begin{tabular}[c]{@{}l@{}}\citet{peng2020faster} (Faster ILOD)$\dagger$\end{tabular}   &  & 40.10  & 20.64         & 0.939                                                             \\
\citet{zhou2020lifelong}$\dagger$                                                        &  & 36.80  & 22.70         & 0.868                                                             \\
\citet{yang2020two}                                                             &  & 43.75  & 24.23         & 0.882                                                             \\
\citet{peng2021sid} (SID)$\dagger$                                                       &  & 41.60  & 25.20         & 0.885                                                             \\
\citet{yang2021multi}                                                           &  & 44.62  & -             & 0.854                                                             \\
\citet{kj2021incremental} (Meta-ILOD)                                           &  & 40.50  & 23.80         & 0.794                                                             \\
\citet{dong2021bridging}$\dagger$                                                        &  & 40.90  & 22.50         & 0.893                                                             \\
\hline
\end{tabular}}
\end{table}

Overall, it is clear that all methods suffered from some aspect of forgetting and were limited to the joint-training baseline in all benchmarks. It is important to mention that this might not always be the case and that some CL strategies may surpass this baseline, which has been the case for some based on parameter freezing mechanisms~\citep{yoon2017lifelong}. The RKT strategy proposed by \citep{ramakrishnan2020relationship} was the best knowledge distillation method on the benchmarks it participated; however, most of the pure distillation-based techniques presented low plasticity probably due to the constraints imposed on the original weights. The IncDet model, which involved EWC regularization and pseudo-labeling, showed the most consistent results in all evaluated benchmarks. Yet, some strategies based on parameter isolation and replay that did well in individual benchmarks, such as MMN and ORE, demonstrated that there is still room for exploring alternatives and possibly combining them. 

This review did not consider other desired characteristics for the CL desiderata, such as the low memory footprint, which usually tends to overthrow parameter isolation strategies and fast adaptability to new categories. The meta-learning hybrid method from ~\citet{kj2021incremental} presented results slightly superior to other knowledge distillation techniques. Regardless, the method quickly adapted to new tasks using only 10 replay samples for each category during fine-tuning. In this way, considering the CL desiderata for object detection discussed in Section~\ref{sec:cl-background}, meta-learning hybrid methods can play an interesting role in class-incremental scenarios and should be more investigated.

\section{Trends and Research Directions}
\label{sec:directions}

Considering the main takeaways of the previous systematic review, in this section, we briefly discuss some of the observed trends and possible research directions in the CIOD field.

\textbf{Hybrid methods prevent more forgetting}: The best performing solutions to the class-incremental problem in object detection involved a combination of techniques to avoid catastrophic forgetting. This outcome agrees with the findings from other computer vision tasks~\citep{qu2021recent} and corroborates with the fact that even the brain has multiple ways to prevent subtle task interference~\citep{hassabis2017neuroscience}. One key point common to most hybrid methods was the fine-tuning on new classes given the representation of old categories using pseudo-labels or replay samples. This fine-tuning resulted in better results but can require a large buffer of samples and, similarly, an extensive hyperparameter search which might prevent its application in the real-world. 

\textbf{Knowledge Distillation: a strong baseline with caveats}: It is easy to notice from Table~\ref{tab-CIOD-papers} that most proposed strategies in the CIOD field use knowledge distillation as their primary mechanism to mitigate the effects of catastrophic forgetting. Comparing the results of the selected papers on the PASCAL VOC 2007 and MS COCO incremental benchmarks considering the metrics that assess the stability-plasticity of solutions, the differences between a recently proposed distillation technique such as~\citet{peng2021sid} and the first work of \citet{shmelkov2017incremental} are subtle. This either means that researchers might have been overfitting their solutions to the benchmarks or that simple logits and bounding box distillation are a strong baseline. We believe the latter to be a more reasonable explanation.

\textbf{Working towards the CL Desiderata for object detection}: The majority of the currently published CL research is done focusing on improving the last 0.01\% of performance, sometimes considering unrealistic scenarios (e.g., use of task labels at test time). However, for real-world focused applications, strategies should also contemplate practical implementation aspects such as the computational burden and frequency of updates for the model. For class-incremental detectors, the desiderata described in Section~\ref{sec:cl-background} give an intuition of the main aspects that future research could focus on in order to increase practical adoption for researchers in academia and industry. 



\textbf{Working towards the standardization of implementations}: Research in CIOD suffers from poor standardization and has not fully adopted the advent already developed by the CL community for reproducibility, such as the Avalanche and Continuum libraries~\citep{lomonaco2021avalanche, douillard2021continuum}. Besides that, there is no standard implementation for most of the discussed solutions to leverage fair comparisons. Although some available implementations are provided using the Detectron2 framework or its old form~\citep{peng2020faster, joseph2021towards, kj2021incremental}, the interpretation of the changes to the original framework that are needed to reach the same results is often difficult due to the abstract structure of its repository. One step towards improving on this issue is the open-sourcing of the code for the regular baselines evaluated when proposing a new benchmark. Ideally, the implementation should envision using a well-established framework specific to the field (e.g., Avalanche), where a better description of the differences and human-readable code can be maintained. Nevertheless, the metrics proposed in this paper are also available as a tool for performing honest comparisons between solutions for the same benchmark.

\textbf{Overcoming the overestimation of results}: As also found in the recent survey for few-shot object detection~\citet{huang2021survey}, most works evaluated on the VOC and COCO datasets are using their training and validation splits for fitting the models and the testing set for selecting hyperparameters. This can lead to an overestimation of results and generalization problems when selecting techniques for good performance in the real world. A straightforward fix would be to use the original train/val/test splits as indicated by the datasets organizers and not perform contradictory actions to favor the proposed methods. Yet, as most researchers are using this setup to report their performance on the current benchmarks, it is sadly expected that the follow-up papers still keep the same choice of splits. We believe researchers should be more careful when proposing and evaluating their strategies for new incremental benchmarks to ensure a not biased outcome. 

\section{Related fields}
\label{sec:related-fields}

Some related computer vision tasks already involve components that deal with incremental object detection in their pipeline. This section discusses a few of them shortly to make it possible for the readers to connect with other fields that can inspire and contribute to the current research of continual object detectors.

\subsection{Open-World Object Detection}

The set of possible objects that a detector can encounter at test time in the wild is limitless. For dealing with the unknown and adapting to it, the field of open-world object detection has emerged as a possible solution to unify the paradigms of open-set and open-world recognition to class-incremental learning with object detectors~\citep{mundt2020wholistic, bendale2015towards}. The solutions in this category are usually the combination of a structure to detect out-of-distribution samples (i.e., unknown objects) and a specific module to allow learning from them in an incremental manner. By modeling the unknown, researchers believe it is possible to reduce the label conflict and therefore enable more autonomous detection pipelines~\citet{joseph2021towards}.

\subsection{Incremental Few-shot object detection}

When learning incrementally in robotics applications, models can be required to learn from data streams with only a few batches and several unseen classes. This scenario makes it difficult to apply the traditional batch learning used with neural networks and therefore needs a particular solution. The field of Incremental Few-Shot Object Detection (iFSD) looks for fast adaption of a trained network in situations of a low-data regime for learning novel classes~\citep{perez2020incremental}. This scenario is naturally more difficult than the plain CIOD paradigm since it assumes no large dataset is provided. Arguably, the current results on their benchmarks show a trend to focus more on the adaption to new classes than on avoiding the forgetting of old ones~\citep{li2021class}. This might indicate that research should be directed first at how to solve a less complicated problem (i.e., class-incremental object detection with large batches), which can give hints on how to move forward to more complex scenarios. 

\subsection{Continual Semantic Segmentation}

The field of Continual Semantic Segmentation deals with the same difficulties of continual object detection (e.g., background label conflict) but at the pixel level. Most of the current solutions that have excelled in the field involve the techniques described in this review, such as pseudo-labeling~\citep{douillard2021plop} and knowledge distillation~\citep{michieli2019incremental, cermelli2020modeling}. Its application has a direct impact on real-world robotics navigation and should always be looked at closely by CIOD researchers for insights. 

\subsection{Zero-shot Object Detection}

The Zero-Shot Object Detection paradigm consists of learning to detect new categories that are not present in the training set by using non-visual features that describe them~\citep{bansal2018zero}. Specifically, a pre-trained language model was originally used to model the semantics associated with the class labels. These relations were then used to guide the learning and inference of new unseen classes by a detector. The method proposed by~\citet{yang2021objects}, which was described in Section~\ref{sec:RQ1}, combined a zero-shot strategy with exemplar replay and showed decent results not only for open-world recognition (their primary goal) but also for some CIOD benchmarks. This might be an indication that innovations can be appropriately adapted and shared among these fields.

\section{Conclusions}
\label{sec:conclusion}

This short systematic review investigated how continual learning solutions have been applied to object detection tasks covering the topic`s technical background up to the most explored benchmarks, metrics, and strategies. 

For the literature review, we analyzed the reported performance of the leading papers in a popular benchmark for the class-incremental scenario with the lens of a new metric explicitly proposed to look at how well a detector adapts and maintains its internal knowledge. We found out that even though most of the current research appeals to the single use of regularization-based techniques, specifically knowledge distillation, the methods that presented the best overall results on the evaluated benchmarks usually combine such techniques with replay, self-labeling, and meta-learning.

Finally, we discussed some of the main trends in the field, pitfalls and how researchers may avoid them, and a few related tasks that can inspire the proposal of new methods and possible future research intersections.

\section*{Acknowledgments}
This study was funded in part by the Coordenação de Aperfeiçoamento de Pessoal de Nível Superior - Brasil (CAPES) - Finance Code 001. The authors also would like to thank the Eldorado Research Institute for supporting this research.



\bibliographystyle{elsarticle-num-names} 
\bibliography{references}

\end{document}